\documentclass{article} % For LaTeX2e
\usepackage{iclr2026_conference,times}
\usepackage{algorithm}
\usepackage{algorithmic}
\usepackage{booktabs}
\usepackage{amsmath}
\usepackage{graphicx}
\usepackage{caption} 
\usepackage{subcaption}
\usepackage{makecell}
\usepackage{titletoc}
\usepackage{amsmath,amsfonts,bm}

\def\eqref#1{equation~\ref{#1}}
\def\1{\bm{1}}

\DeclareMathAlphabet{\mathsfit}{\encodingdefault}{\sfdefault}{m}{sl}
\SetMathAlphabet{\mathsfit}{bold}{\encodingdefault}{\sfdefault}{bx}{n}

\usepackage{multirow}
\usepackage{hyperref}
\usepackage{url}

\title{Geometry Forcing: Marrying Video Diffusion and 3D Representation for Consistent World Modeling}

\author{
\textbf{Haoyu Wu}$^{1}$\thanks{Equal contribution.} , 
\textbf{Diankun Wu}$^{2}$\footnotemark[1], 
\textbf{Tianyu He}$^{1}$\thanks{Project lead.}, 
\textbf{Junliang Guo}$^{1}$, 
\textbf{Yang Ye}$^{1}$, \\
\textbf{Yueqi Duan}$^{2}$, 
\textbf{Jiang Bian}$^{1}$ \\
\\
$^1$Microsoft Research \quad 
$^2$Tsinghua University \quad 
}

\iclrfinalcopy % Uncomment for camera-ready version, but NOT for submission.

\begin{document}

\maketitle

\begin{abstract}

Videos inherently represent 2D projections of a dynamic 3D world. However, our analysis suggests that video diffusion models trained solely on raw video data often fail to capture meaningful geometric-aware structure in their learned representations. To bridge the gap between video diffusion models and the underlying 3D nature of the physical world, we propose Geometry Forcing, a simple yet effective method that encourages video diffusion models to internalize 3D representations. Our key insight is to guide the model’s intermediate representations toward geometry-aware structure by aligning them with features from a geometric foundation model. To this end, we introduce two complementary alignment objectives: Angular Alignment, which enforces directional consistency via cosine similarity, and Scale Alignment, which preserves scale-related information by regressing geometric features from normalized diffusion representations. We evaluate Geometry Forcing on both camera-view conditioned and action-conditioned video generation tasks. Experimental results demonstrate that our method substantially improves visual quality and 3D consistency over the baseline methods.

Project page: \href{https://GeometryForcing.github.io}{https://GeometryForcing.github.io}

\end{abstract}

\begin{figure}[htbp]
    \centering
    \includegraphics[
        width=0.95\linewidth,
        clip
    ]{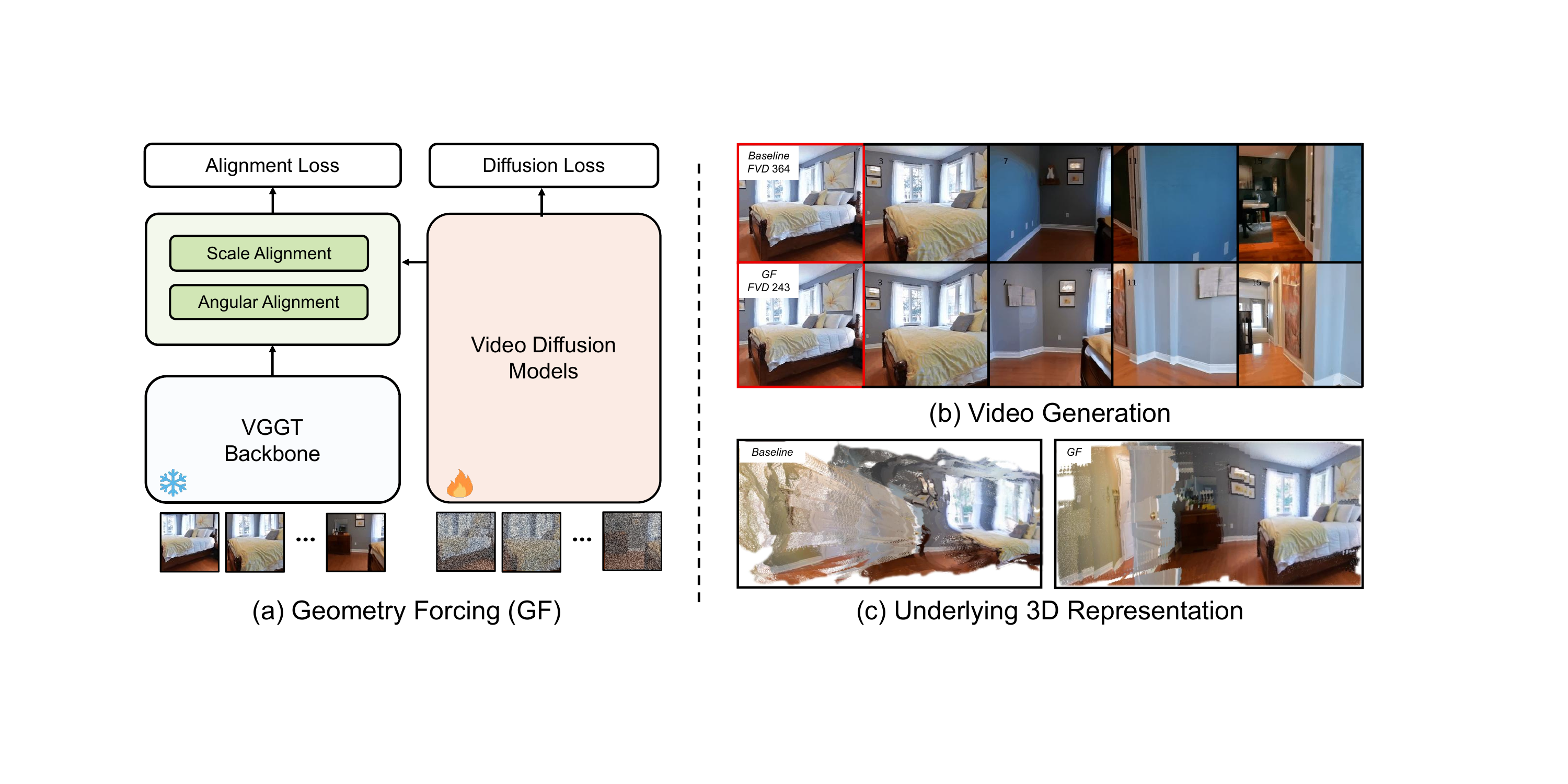}
    \caption{
    \textbf{Geometry Forcing equips video diffusion models with 3D awareness.} 
    \textbf{(a)} We propose Geometry Forcing (GF), a simple yet effective paradigm to internalize geometric-aware structure into video diffusion models by aligning with features from a geometric foundation model, \emph{i.e.}, VGGT~\citep{wang2025vggt}. 
    \textbf{(b)} Compared to the baseline method~\citep{dfot}, our method produces more consistent generations both temporally and geometrically. 
    \textbf{(c)} Features learned by the baseline model fail to reconstruct meaningful 3D geometry, whereas our method internalizes 3D representation, enabling accurate 3D reconstruction from the intermediate features.
    }
    \label{fig:head}
\end{figure}

\section{Introduction}
\label{sec:intro}

Learning to simulate the physical world and predict future states is a cornerstone of intelligent systems~\citep{ha2018world}. Recent advances in generative modeling~\citep{ho2020denoising,rombach2022high,peebles2023scalable,brown2020language}, coupled with the availability of large-scale video datasets, have led to significant progress in generating realistic visual environments conditioned on text descriptions~\citep{sora,yang2024cogvideox,polyak2024movie,veo3} or agent actions~\citep{hu2023gaia,guo2025mineworld,bar2025navigation_NVM}. However, these approaches typically aim to model pixel distributions across video frames, overlooking a fundamental principle: \emph{videos are 2D projections of a dynamic 3D world}~\citep{glassner1989introduction}. By focusing solely on image-space generation, such models often struggle to maintain geometric coherence and long-term consistency, particularly in autoregressive settings where small errors can accumulate over time~\citep{dforcing,cheng2025nfd,huang2025selforcing}.

Building on this motivation, a growing line of research has explored the explicit modeling of the dynamic 3D structure of the physical world~\citep{zhu2024compositional,team2025aether,jiang2025geo4d}, rather than implicitly learning distributions in 2D pixel space. For example, \citet{zhang2025world} proposes transforming 3D coordinates into point maps and jointly modeling the RGB and 3D information. While effective to some extent, representing 3D information in a tractable form remains challenging, and reliance on additional annotations limits scalability.

In this work, we aim to bridge the gap between video diffusion models and the underlying dynamic 3D structure of the physical world. We begin with a fundamental question: \emph{Can video diffusion models implicitly learn 3D information through training on raw video data, without explicit 3D supervision}? To investigate this, we analyze a pretrained video diffusion model~\citep{dfot} by introducing a DPT~\citep{ranftl2021vision} head that maps its intermediate features to corresponding depth maps~\citep{wang2025vggt}. As illustrated in Fig.~\ref{fig:head}(c), we observe that features learned solely from raw video data fail to yield meaningful geometric representations, highlighting a potential gap in the geometric understanding of video diffusion models trained without additional guidance.

To address this limitation, we propose \emph{Geometry Forcing~(GF)}, a simple yet effective approach that encourages video diffusion models to \emph{internalize} 3D representations during training. Inspired by recent advances in semantic REPresentation Alignment (REPA) for image diffusion models~\citep{yu2024representation}, we align features of video diffusion models with the \emph{geometric representations} from a pretrained 3D foundation model~\citep{wang2025vggt}. To align these two representations, our method introduces two complementary alignment objectives: Angular Alignment and Scale Alignment. Angular Alignment enforces directional consistency between the diffusion model’s features and geometric representations by maximizing their cosine similarity. Scale Alignment, in contrast, preserves the scale information of the geometric representations by predicting geometric features from normalized diffusion features. The decoupled formulation of Angular and Scale Alignment allows the model to capture both directional and scale-related aspects of geometry, while improving stability during training and expressiveness in the learned representations.

We evaluate the effectiveness of GF on two widely adopted benchmarks: camera-view-conditioned video generation on RealEstate10K~\citep{zhou2018stereo} and action-conditioned video generation in the Minecraft environment~\citep{baker2022video}. Experimental results demonstrate that our method delivers substantial gains in geometric consistency and visual quality over the baseline methods. For example, GF reduces the FVD from 364 to 243 on the RealEstate10K benchmark. Moreover, the ability to reconstruct explicit geometry during inference opens opportunities to integrate structured memory into long-term world modeling.

\section{Related Work}

\subsection{Interactive World Modeling}

A world simulator seeks to model the underlying dynamics of the physical world by predicting future states conditioned on current observations and conditions. We review prior works through the lenses of interactive video generation, 4D generation, and consistent world modeling.

\paragraph{Interactive Video Generation.}

Recent advancements in generative models~\citep{ho2020denoising,rombach2022high,peebles2023scalable,lipman2023flow,bruce2024genie,parker2024genie2,alonso2024atari,valevski2024diffusion} have positioned video generation as a promising approach to world modeling. Beyond text-to-video synthesis~\citep{videocrafter1,chen2024videocrafter2,kong2024hunyuanvideo,wan2025wan,li2024t2v,liu2025videodpo,ye2025fast}, interactive video generation~\citep{IGVsurvey} that emphasizes responding interactive control signals evolves rapidly. Existing models incorporate different signals like camera controls~\citep{he2024cameractrl,Yu2024ViewCrafterTV,dfot} and action controls~\citep{oasis2024,guo2025mineworld,feng2024matrix,shin2024wham}. Building on this progress, our work introduces a novel training pipeline that enhances 3D consistency in video generation, enabling more coherent and realistic simulation of spatial scenes.

\paragraph{Interactive 4D Generation.}

In contrast to data-driven video simulators, 4D-based simulators~\citep{chung2023luciddreamer,4d-fy,Wu_2025_CVPR,yu2025wonderworld,lee2024vividdream} explicitly model dynamic 3D structures ~\citep{kerbl20233d,mildenhall2021nerf,xiang2025TRELLIS}. Building upon 3D content generation~\citep{raj2023dreambooth3d}, these methods evolve from dynamic objects~\citep{xu2024comp4d,bahmani2024tc4d} to complex dynamic scenes~\citep{niemeyer2021giraffe,zhu2024compositional}. Recent works integrate video priors to improve the realism and temporal coherence of 4D~\citep{team2025aether,jiang2025geo4d,mai2025can,chen2025flexworld}. For example, TesserAct~\citep{zhen2025tesseract} predicts RGB, depth, and surface normals to reconstruct temporally consistent 4D scenes. While our work shares the goal of unifying 3D and video generation, it differs in that it injects 3D geometric priors into the video representation to improve temporal and spatial coherence.

\paragraph{Consistent World Modeling.}
A key challenge in world modeling lies in maintaining consistency over long video sequences. To address this, prior works have explored different solutions. Frame-level context mechanisms~\citep{dforcing,fuest2025maskflow,mambaworldmodel,videospm} improve consistency by training with noisy context frames. Meanwhile, other methods leverage 3D information. For example, \citet{xiao2025worldmem} maintains a memory bank indexed by field-of-view overlap to retrieve relevant historical frames. \citet{zhang2025world} proposes jointly modeling RGB frames and point maps to maintain 3D consistency. In contrast, we propose directly incorporating 3D representations into video diffusion models, thereby enabling more stable geometric consistency.

\subsection{3D Foundation Models}
3D foundation models (3DFMs)~\citep{li2024megasam,yang2025fast3r,smart2024splatt3r,wang2025cut3r,wang2024dust3r} have recently shown remarkable progress, applying end-to-end framework with fast and robust inference. These models are capable of predicting different 3D properties, including camera poses~\citep{zhang2025flare}, depth maps~\citep{piccinelli2024unidepth}, and dense point clouds~\citep{wang2025vggt}, directly from visual inputs.

Due to their accuracy, efficiency, and robustness, 3DFMs are becoming essential for enabling downstream tasks like spatial reasoning~\citep{wu2025spatial,huang2025mllms,fan2025vlm3r}, autonomous driving~\citep{Fei2024Driv3RLD}, SLAM~\citep{liu2025slam3r, Maggio2025VGGTSLAMDR}, and beyond. Inspired by their strong 3D capabilities, we explore incorporating 3D representations into video diffusion models to enhance temporal and spatial consistency for world modeling.

\section{Preliminaries}
\label{sec:preliminaries}

Our approach builds upon autoregressive video diffusion models~\citep{dforcing,dfot,cheng2025nfd} and incorporates a 3D foundation model~\citep{wang2025vggt} into the training process to guide geometric learning. In this section, we provide a brief overview of both components to establish the foundation for our method.

\subsection{Autoregressive Video Diffusion Models}
\label{sec:arvdm}

\paragraph{Training.}

We formulate the training pipeline based on Flow Matching~\citep{lipman2023flow,liu2023flow} with a Transformer backbone~\citep{vaswani2017attention,bao2023uvit} to achieve simplicity and scalability. Let $\mathbf{x} = \{x_1, \ldots, x_I\}$ denote a video sequence sampled from the data distribution. We assign an independent timestep for each frame $\mathbf{t} = \{t_1, \ldots, t_I\}$ and corrupt frames via interpolation:
\[
x_i^{t_i} = (1 - {t_i}) \cdot x_i^0 + {t_i} \cdot \epsilon_i, \quad \text{where} \quad \epsilon_i \sim \mathcal{N}(0, I).  % $v_i^{t_i} \equiv \epsilon_i - x_i^{t_i}$
\]
The target velocity field is defined as the difference between the noise and the clean input. We train a neural network $v_\theta$ to minimize the Flow Matching loss:
\[
\mathcal{L}_{\text{FM}} = \left\|  v_\theta ( \mathbf{x}^\mathbf{t}, \mathbf{t}) - (\boldsymbol{\epsilon} - \mathbf{x}) \right\|^2.
\]

\paragraph{Sampling.}

At inference time, the sampling follows a simple probability flow ODE:
\[
\text{d}\mathbf{x} = v_\theta ( \mathbf{x}^\mathbf{t}, \mathbf{t}) \cdot \text{d}\mathbf{t}.
\]
In practice, we iteratively apply the standard Euler solver~\citep{euler1845institutionum} to sample data from noise. For autoregressive generation, we initialize the inputs with a clean context and generate subsequent frames sequentially, conditioning each prediction on the previously generated frames.

\subsection{Visual Geometry Grounded Transformer}
\label{sec:vggt}

Visual Geometry Grounded Transformer (VGGT)~\citep{wang2025vggt} is a feed-forward model that directly outputs 3D attributes of a scene, including camera parameters, point maps, and depth maps.

VGGT has a Transformer backbone and multiple prediction heads. The model employs an Alternating-Attention mechanism that interleaves frame-wise self-attention and global self-attention to extract local and global information. For each frame, local and global features are integrated into a unified representation, which is subsequently processed by task-specific heads to produce 3D attributes. We leverage the features from the Transformer backbone of VGGT to extract geometric representation.

\section{Geometry Forcing}
\label{sec:method}

\subsection{Method Overview}

\paragraph{Motivation.}

Recent advances in video diffusion models have enabled the simulation of the world directly from large-scale video datasets. However, these models often overlook a fundamental property of visual data: videos are 2D projections of a dynamic 3D world. To address this, we seek to narrow the gap between video diffusion models and the dynamic 3D structure of the world.
\vspace{-0.5em}
\paragraph{Observation.}

We begin by examining whether video diffusion models are capable of implicitly learning 3D information when trained solely on raw video data, without explicit 3D supervision. To probe the geometric content of their learned representations, we adopt a strategy inspired by linear probing~\citep{he2020momentum}: we freeze the parameters of a pretrained video diffusion model~\citep{dfot} and train a DPT~\citep{ranftl2021vision} head to map intermediate features to corresponding depth maps~\citep{wang2025vggt}. This allows us to assess the extent to which geometric information is encoded in the model's feature space. The results, presented in Fig.~\ref{fig:head}(c), indicate that features learned solely from raw video data do not produce meaningful geometric representations, suggesting a limited capacity of the model to encode dynamic 3D structure without explicit geometric guidance.
\vspace{-0.5em}
\paragraph{Challenge.}

Bridging the gap between video diffusion models and the dynamic 3D structure of the world presents significant challenges, primarily due to the limited annotated 3D data. A straightforward approach is to jointly model RGB and geometric information within an end-to-end architecture. However, relying heavily on 3D annotations can hinder the scalability and generalization of the models, particularly when applied to large, diverse real-world video datasets. 

In this work, inspired by recent advances in REPA~\citep{yu2024representation}, we propose \emph{Geometry Forcing~(GF)} that aligns the features of video diffusion models with geometric representations, encouraging the model to internalize geometric information. Our approach builds upon video diffusion models described in Sec.~\ref {sec:arvdm}. In Sec.~\ref {sec:alignment}, we introduce two regularization objectives designed to facilitate representation alignment between the diffusion model and geometric foundation model. The overall training objective, along with additional functional extensions, is summarized in Sec.~\ref {sec:overall}.

\subsection{Geometric Representation Alignment}
\label{sec:alignment}

To improve the geometric consistency of the learned representations, we introduce two complementary alignment objectives: \emph{Angular Alignment} and \emph{Scale Alignment}. These objectives are designed to align the latent features of the diffusion model with intermediate representations from a pretrained geometric foundation model~\citep{wang2025vggt}, ensuring both directional consistency and scale preservation of geometric features within the feature space.

\paragraph{Angular Alignment.}
\label{sec:angular}

Angular Alignment enforces directional correspondence between the hidden states of the diffusion model, denoted by $h$, and specified target features, denoted by $y$. We select intermediate features from the Transformer backbone of VGGT~\citep{wang2025vggt} as $y$, as these features preserve both local and global information within each frame and can be further used to reconstruct various explicit geometric representations. In practice, the target features \(y \in \mathbb{R}^{L \times N \times P \times D}\),  where \(L\) denotes the number of layers, \(N\) denotes the number of input images, \(P\) denotes the patch count, and \(D\) denotes the feature dimension. To achieve Angular Alignment, we first use a lightweight projector \(f_\phi\) to map the diffusion latents \(h \in \mathbb{R}^{N \times P' \times D'}\) to \(y\)'s shape. The Angular Alignment loss is then defined as:
\[
\mathcal{L}_{\text{Angular}} = - \frac{1}{LNP} \sum_{\ell=1}^{L} \sum_{n=1}^{N} \sum_{p=1}^{P} \cos\left( y_{\ell,n,p},\; f_\phi(h_{n,p}) \right),
\]
where \(\cos(\cdot,\cdot)\) denotes cosine similarity. This loss aligns hidden states independently at both the frame and patch levels. Since the VGGT backbone already incorporates cross-frame attention, we do not explicitly enforce global alignment across frames in the loss.

\paragraph{Scale Alignment.}
\label{sec:scale}

While Angular Alignment ensures directional consistency, it disregards feature scale, which can also encode geometric information. Although direct mean squared error (MSE) loss could supervise magnitudes, it often leads to optimization instability and model collapse due to inherent scale differences across models. To address this issue, we introduce Scale Alignment, which preserves scale information through predicting the scale of target features given normalized diffusion hidden states. Specifically, we first normalize \(f_\phi(h)\) to unit length. Then we use another lightweight prediction head \(g_\varphi\) to predict the full target features from normalized inputs:
\[
\hat{h}_{\ell,n,p} = \frac{f_\phi(h_{n,p})}{\|f_\phi(h_{n,p})\|_2}, \quad \tilde{y}_{\ell,n,p} = g_\varphi(\hat{h}_{\ell,n,p}).
\]
The Scale Alignment loss is defined as:
\[
\mathcal{L}_{\text{Scale}} = \frac{1}{LNP} \sum_{\ell=1}^{L} \sum_{n=1}^{N} \sum_{p=1}^{P} \left\| \tilde{y}_{\ell,n,p} - y_{\ell,n,p} \right\|_2^2.
\]
This decomposition stabilizes training while capturing both directional and scale attributes of geometric representations.

\subsection{3D-aware Autoregressive Video Diffusion Models}
\label{sec:overall}

Building on the autoregressive video diffusion framework and the proposed alignment objectives, we now present the overall training objective:
\[
\mathcal{L} = \mathcal{L}_{\text{FM}} + \lambda_{\text{Angular}} \cdot \mathcal{L}_{\text{Angular}} + \lambda_{\text{Scale}} \cdot \mathcal{L}_{\text{Scale}}.
\]
Given the intermediate features of our model are well-aligned with geometric representations, an appealing consequence is the model’s ability to predict explicit 3D geometry during inference. This enables unified generation of both video and 4D, effectively bridging the gap between videos and the underlying dynamic 3D structure of the physical world, as illustrated in Fig.~\ref{fig:head}. Moreover, the ability to reconstruct explicit geometry during inference provides a structured and interpretable form of memory that can be further used to support long-term world modeling. We leave the exploration of such geometry-based memory mechanisms as a promising direction for future work. 

\paragraph{Discussion.}

Teacher Forcing~\citep{williams1989learning} is a widely adopted training paradigm for autoregressive models~\citep{radford2019language,brown2020language,kondratyuk2024videopoet}. To combine the autoregressive nature with diffusion models, Diffusion Forcing~\citep{dforcing} proposes training video diffusion models with independent noise levels for each frame. More recently, Self Forcing~\citep{huang2025selforcing} has been proposed to address exposure bias in autoregressive video diffusion models. Orthogonal to these methods, Geometry Forcing focuses on improving the structure of the learned representations by aligning the intermediate representation of video diffusion models with geometry-aware signals from a 3D foundation model. Our approach provides structural supervision at the representation level, encouraging the model to internalize 3D consistency throughout training.

\section{Experiments}

\vspace{-0.5em}
In this section, we evaluate Geometry Forcing~(GF) on camera-view-conditioned video generation on the RealEstate10K~\citep{zhou2018stereo} dataset and action-conditioned video generation on the Minecraft environment~\citep{baker2022video}. We provide additional illustrations and visualizations in the Appendix.

\vspace{-0.5em}
\paragraph{Implementation Details.}  

For camera view-conditioned video generation, we apply GF on Diffusion Forcing Transformer~\citep{dfot}, training on 16-frame 256\(\times\)256 videos for 2,500 steps with a learning rate of \(8 \times 10^{-6}\) and batch size 8. Inference is conditioned on the first frame and per-frame camera poses. For action-conditioned video generation, we apply GF to Next-Frame Diffusion~\citep{cheng2025nfd}, training on 32-frame 384\(\times\)224 videos for 2,000 steps with a learning rate of \(6 \times 10^{-5}\) and batch size 32. We set $\lambda_{\text{Angular}} = 0.5$ and $\lambda_{\text{Scale}} = 0.05$ to balance each loss component. All experiments are conducted on 8 NVIDIA A100 GPUs. 

\vspace{-0.5em}
\paragraph{Evaluation Metrics.}  
\label{sec:metrics}

We evaluate visual quality using FVD (Fréchet Video Distance)~\citep{unterthiner2018towards}, PSNR (Peak Signal-to-Noise Ratio), SSIM (Structural Similarity Index)~\citep{wang2004image}, and LPIPS (Learned Perceptual Image Patch Similarity)~\citep{zhang2018unreasonable}. 

To further evaluate geometric consistency, we introduce Reprojection Error (RPE)~\citep{duan2025worldscore} and Revisit Error (RVE)~\citep{xiao2025worldmem}. Reprojection Error (RPE) quantitatively measures multi-view geometric consistency by calculating the average reprojection discrepancy between projected and observed pixel locations across frames. Revisit Error (RVE) assesses long-range temporal consistency by examining discrepancies between initial and revisited frames under complete camera rotation. We provide more details of these metrics in the Appendix (Sec.~\ref {supp:reproj}).

\begin{figure}[t]
    \centering
    \includegraphics[
        width=0.95\linewidth,
        clip
    ]{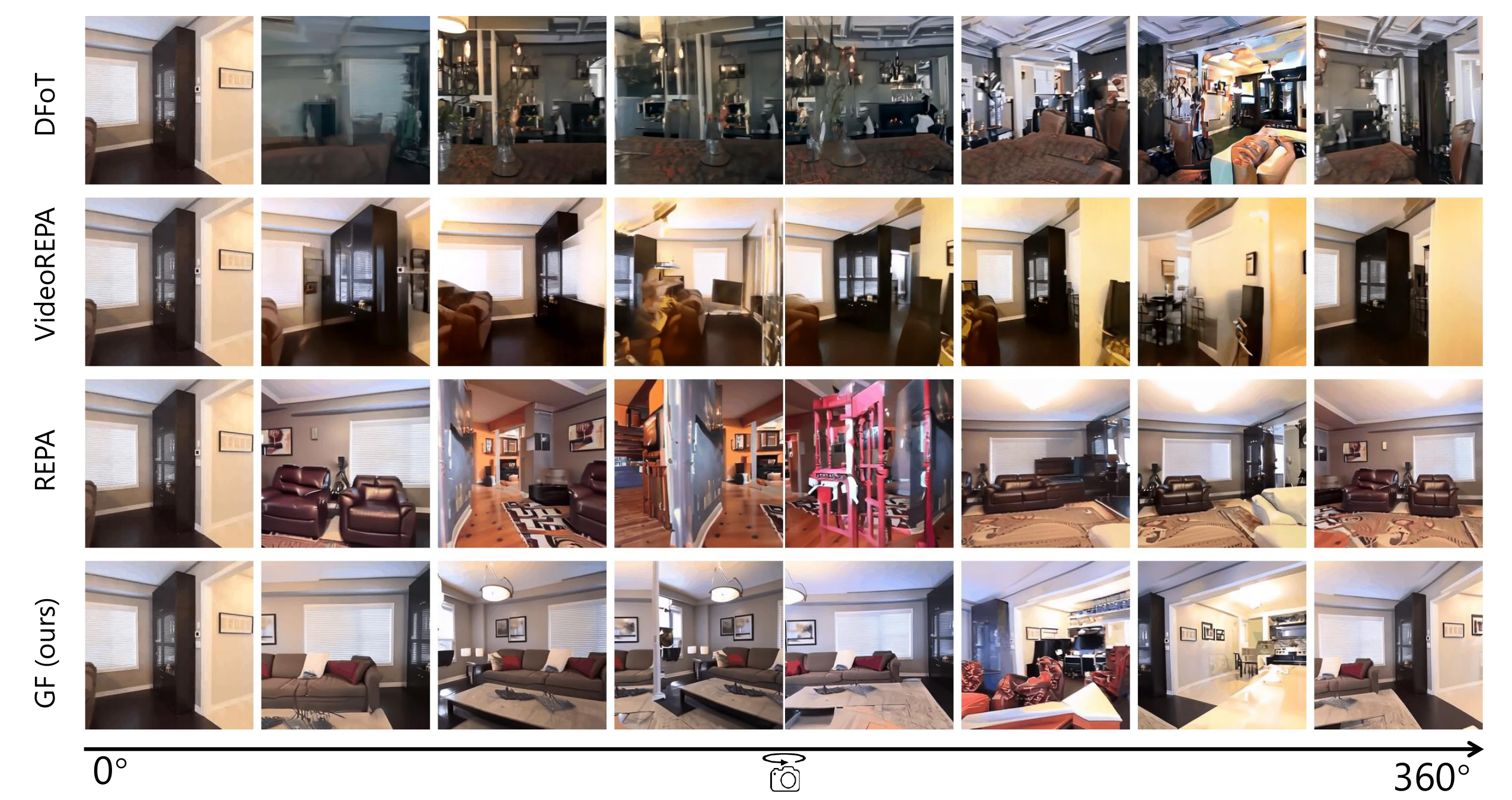}
    \caption{\textbf{Qualitative comparison of camera view-conditioned video generation under full-circle rotation.}  Videos are generated from a single frame, and per-frame camera poses simulate a full 360° rotation.  Our method (GF) is compared with DFoT~\citep{dfot}, VideoREPA~\citep{zhang2025videorepa}, and REPA~\citep{yu2024representation}. The results demonstrate that the baseline methods fail to maintain temporal consistency, while our proposed GF consistently revisits the starting viewpoint.}
    \label{fig:rotate}
\end{figure}

\begin{table}[t]
\centering
\caption{
Quantitative comparison on the RealEstate10K dataset for both short-term (16-Frame) and long-term (256-Frame) video generation. 
Geometry Forcing substantially improves over the baseline. \textbf{bold} values denote the best, and \underline{Underlined} values indicate the second best.  \textbf{*} indicates the method is conditioned on the first frame only.}
\label{tab:re10k_results}
\small
\resizebox{\textwidth}{!}{
\begin{tabular}{lccccccc}
\toprule
\textbf{Method} & \textbf{Frames} & \textbf{FVD↓} & \textbf{LPIPS↓} & \textbf{SSIM↑} & \textbf{PSNR↑} & \textbf{RPE↓} & \textbf{RVE↓} \\
\midrule
DFoT~\citep{dfot}       & 16   & 252  & 0.40 & 0.50 & 14.40 & --    & --    \\
REPA~\citep{yu2024representation}         & 16   & 221  & 0.37 & 0.54 & \textbf{15.20} & --    & --    \\
VideoREPA~\citep{zhang2025videorepa}     & 16   & 210  & 0.37 & 0.54 & \textbf{15.20} & --    & --    \\
Geometry Forcing (ours)     & 16   & \underline{193}  & \textbf{0.32} & \textbf{0.58} & \underline{14.70} & --    & --    \\
Geometry Forcing (ours) + REPA     & 16   & \textbf{179}  & \underline{0.34} & \underline{0.54} & \underline{15.00} & --    & --    \\
\midrule
Cosmos*~\citep{agarwal2025cosmos}              & 256   & 934   & 0.68   & 0.20   & 10.25    & --    & --    \\
DFoT~\citep{dfot}      & 256  & 364  & 0.55 & 0.36 & 11.40 & 0.3575    & 297    \\
REPA~\citep{yu2024representation}         & 256  & 297  & 0.54 & 0.36 & 11.51 & \underline{0.3337}   & 315    \\
VideoREPA~\citep{zhang2025videorepa}     & 256  & 455  & 0.56 & 0.35 & 11.50 & 0.3823    & \textbf{190}    \\
Geometry Forcing (ours)     & 256  & \underline{243}  & \textbf{0.51} & \textbf{0.38} & \underline{11.87} & \underline{0.3337}   & 272    \\
Geometry Forcing (ours) + REPA     & 256  & \textbf{237}  & \textbf{0.51} & \underline{0.37} & \textbf{12.10} & \textbf{0.3264 } & \underline{236}    \\
\bottomrule
\end{tabular}
}
\end{table}

\subsection{Main Results}
\vspace{-0.5em}
This section presents the main experimental results, comparing our method against state-of-the-art approaches across different tasks. The evaluation results demonstrate the effectiveness and generalization ability of our method in both short- and long-term video generation.

\vspace{-0.5em}
\paragraph{Camera view-conditioned Video Generation.}

We conduct a comprehensive evaluation of GF on the RealEstate10K~\citep{zhou2018stereo} dataset, comparing against state-of-the-art baselines. We report results for both short-term (16-Frame) and long-term (256-Frame) video generation in Tab.~\ref{tab:re10k_results}.

\vspace{-0.5em}
As shown in Tab.~\ref{tab:re10k_results}, our method consistently outperforms all baselines across multiple evaluation metrics, including FVD, LPIPS, SSIM, and PSNR, in both the short-term and long-term generation settings. These results highlight the effectiveness of GF in enhancing visual fidelity, temporal stability, and 3D spatial consistency, thereby enabling more realistic and coherent world modeling. 

\vspace{-0.5em}
\paragraph{Action-conditioned Video Generation.} To demonstrate the generality of our method, we apply GF to Next-Frame Diffusion~\citep{cheng2025nfd} model. As shown in Tab.~\ref{tab:nfd_fvd_only}, the model achieves a lower FVD score, indicating that GF can be seamlessly integrated into video diffusion models and yields measurable gains. Note that there exists a large data distribution gap between the real world and Minecraft. These results demonstrate that GF generalizes well on out-of-domain distributions.

\vspace{-0.5em}
\subsection{Qualitative Results}
\vspace{-0.5em}
Fig.~\ref{fig:rotate} presents qualitative comparisons on the RealEstate10K dataset. Each video is generated from a single input frame along with per-frame camera poses simulating a full 360° rotation. We compare GF against three strong baselines: DFoT~\citep{dfot}, REPA~\citep{yu2024representation}, and VideoREPA~\citep{zhang2025videorepa}. As shown in Fig.~\ref{fig:rotate}, our method reconstructs the initial frame upon completion of the camera rotation, while producing reasonable and realistic intermediate views. In contrast, the baseline methods fail to maintain temporal coherence and scene consistency, resulting in implausible intermediate frames and an inability to revisit the starting viewpoint. These results highlight the superior long-term 3D consistency and scene understanding of our approach.

\begin{table}[t]
\centering
\begin{minipage}[t]{0.48\textwidth}
  \centering
  % \begin{table}[h]
\centering
\caption{\textbf{Ablation study on target representation}. We compare the effect of aligning the diffusion model with different target representations:
DINOv2 (semantic), VGGT (geometric), and their combination.
The joint use of both representation achieves the best FVD.
}
\small
\begin{tabular}{l c}
\toprule
\textbf{Target Representation} & \textbf{FVD-256} \\
\midrule
Baseline & 364 \\
DINOv2 Only & 297 \\
VGGT Only & 243 \\
VGGT + DINOv2 & \textbf{237} \\
\bottomrule
\end{tabular}
\label{tab:target_representation}
% \end{table}

\end{minipage}%
\hfill
\begin{minipage}[t]{0.48\textwidth}
  \centering
  
\caption{\textbf{Ablation study on alignment loss.} Angular and Scale Alignment losses are evaluated for long-term video generation, with MSE as a naive baseline of aligning both angular and scale information. The combination of Angular and Scale Alignment yields the best results.}
\centering
\begin{tabular}{p{3.9cm} c}
\toprule
\textbf{Alignment Loss} & \textbf{FVD-256} \\
\midrule
Baseline & 364.0 \\
Angular & 253.0 \\
Angular + Scale & \textbf{243.0} \\
MSE & 1648.0 \\
\bottomrule
\end{tabular}
\label{tab:loss_ablation}

\end{minipage}
\end{table}

\vspace{-0.5em}
\subsection{Ablation Studies}

\vspace{-0.3em}
We provide a series of ablation studies to validate the design of GF.

\vspace{-0.5em}
\paragraph{Which Representation Should be Aligned?}

To validate the effectiveness of geometric representation, we compare two target representations in GF: VGGT~\citep{wang2025vggt}, trained on 3D datasets with strong geometric priors, and DINOv2~\citep{oquab2023dinov2}, trained on 2D images focusing on semantic features. As shown in Tab.~\ref{tab:target_representation}, aligning with VGGT consistently outperforms DINOv2 on both long-term and short-term generation tasks, highlighting the advantage of geometric alignment over semantic supervision.

To further explore their complementarity, we combine VGGT and DINOv2 features as joint supervision targets. Results in Tab.~\ref{tab:target_representation} show that integrating geometric and semantic signals leads to additional gains, suggesting that the two types of representations are orthogonal and can enhance each other when used together.  However, as we mainly focus on bridging the gap between the video diffusion model and the dynamic 3D structure of the real world, we use only VGGT features in subsequent experiments.

\vspace{-0.5em}
\paragraph{Alignment Loss.} 

GF consists of two alignment objectives: Angular Alignment and Scale Alignment. To validate their effectiveness, we compare three alignment loss types: (1) Angular Alignment alone (Sec.~\ref{sec:angular}), (2) Angular Alignment with Scale Alignment (Sec.~\ref{sec:scale}), and (3) MSE loss between VGGT and diffusion features. As shown in Tab.~\ref{tab:loss_ablation}, the combination of Angular Alignment and Scale Alignment achieves the best performance, indicating the benefit of aligning both angular and scale-related information. Although direct mean squared error (MSE) also supervises magnitudes, changes in the diffusion model's feature scale may cause collapse in the following layers. These results highlight that neither Angular Alignment nor Scale Alignment alone is sufficient.

\begin{table}[t]
\centering
\begin{minipage}[t]{0.48\textwidth}
  \centering
  
  \caption{\textbf{Ablation study on explicit and implicit geometry information.}
We compare the explicit geometry condition with internal alignment (ours).
}
\label{tab:controlnet_repa}
\begin{tabular}{lc}
\toprule
\textbf{Method} & \textbf{FVD-256↓} \\
\midrule
Baseline & 364 \\
Explicit geometry & 280\\
Geometry Forcing (ours)  & \textbf{243} \\
\bottomrule
\end{tabular}
\end{minipage}%
\hfill
\begin{minipage}[t]{0.48\textwidth}
  \centering
  % \begin{table}[h]
% \centering
\caption{\textbf{Evaluation on action-conditioned video generation in Minecraft.}
FVD results of NFD before and after applying Geometry Forcing (GF) on 16-Frame generation show clear improvement.}
\small
\begin{tabular}{l c}
\toprule
\textbf{Method} & \textbf{FVD-16↓} \\
\midrule
NFD            & 216 \\
NFD + GF & \textbf{205} \\
\bottomrule
\end{tabular}
\label{tab:nfd_fvd_only}
% \end{table}

\end{minipage}
\end{table}

\vspace{-0.5em}
\paragraph{Explicitly or Implicitly Integrate Geometry Information into Video Diffusion Models?}

To assess the benefit of internalizing geometric representations, we compare two ways of incorporating geometry into the video diffusion model: internal alignment via GF and external guidance via a ControlNet~\citep{controlnet}. For external guidance, we reconstruct the 3D scene, render it into 2D images, and inject the rendered images as geometric conditions. As shown in Tab.~\ref{tab:controlnet_repa}, using the same VGGT features, GF outperforms rendered-image conditioning. The result shows that while explicit geometric cues are helpful, internal alignment through GF provides consistently stronger supervision. By aligning internal features with geometric representations, GF enables a deeper geometric understanding and yields better performance in perceptual quality and structural consistency. Full evaluation results are listed in Tab.~\ref{tab:explicit_control}. 

\vspace{-0.5em}
\paragraph{Which Layer Should be Aligned?}  

\begin{figure}[t]
    \centering
    \begin{minipage}[t]{0.45\linewidth}
        \centering
        \includegraphics[width=\linewidth]{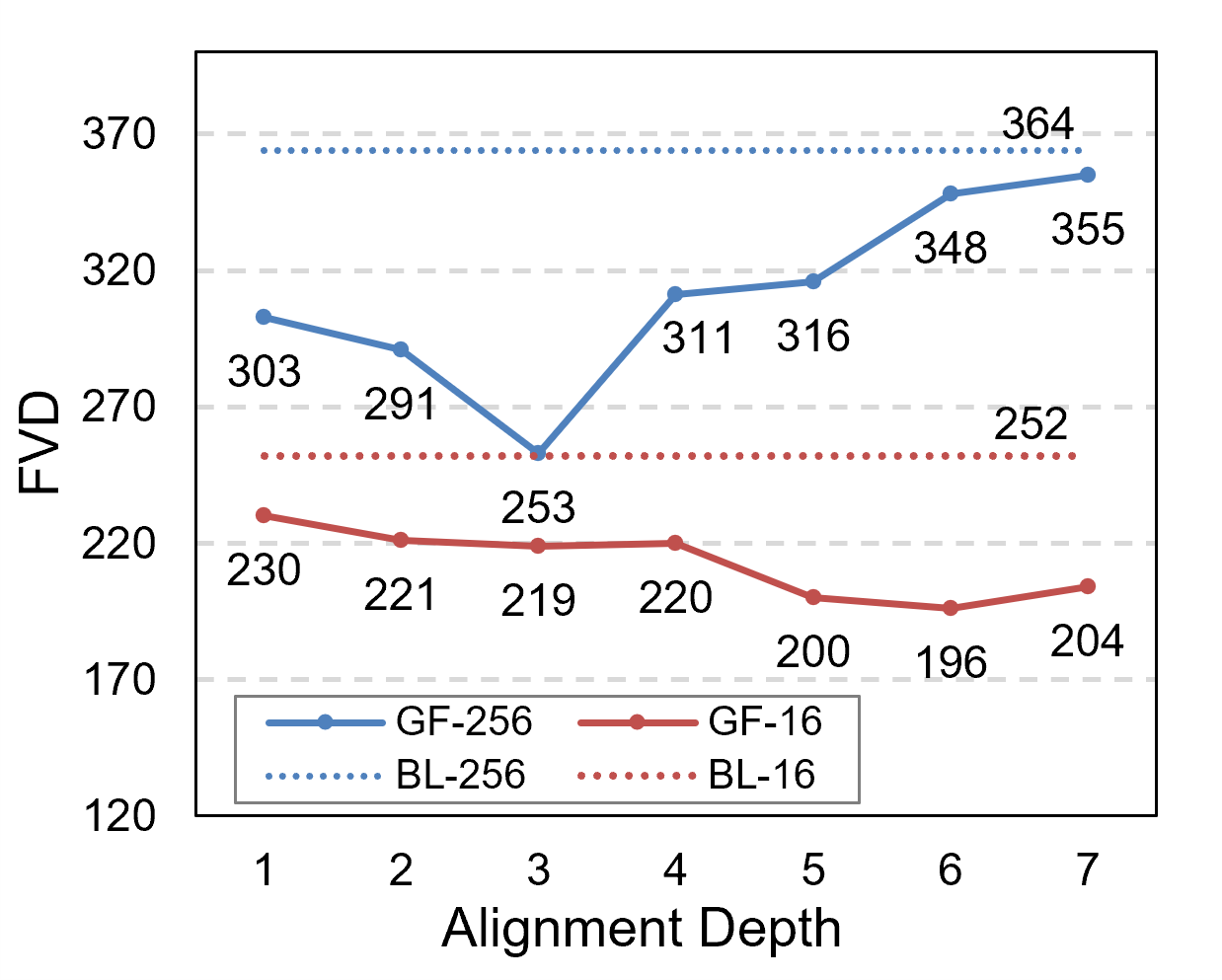}
        \vspace{-2em}
        \caption{\textbf{Ablation study on alignment depth.} We present FVD-256 and FVD-16 results for different alignment layers of the diffusion model, which suggest that mid-level features are most effective to improve video quality.}
        \label{fig:alignment_fvd_depth}
    \end{minipage}
    \hfill
    \begin{minipage}[t]{0.45\linewidth}
        \centering
        \includegraphics[width=\linewidth]{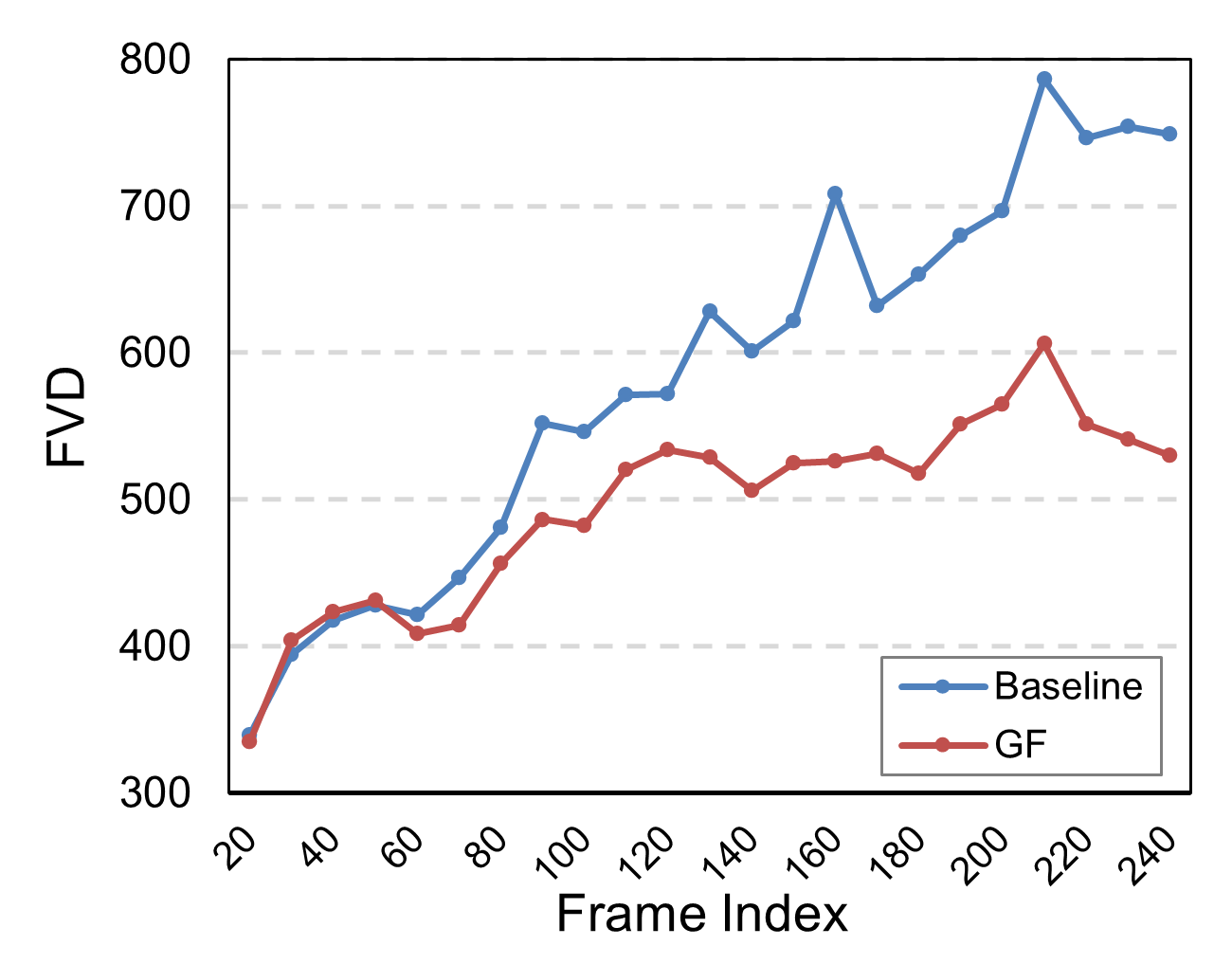}
        \vspace{-2em}
        \caption{\textbf{Exposure bias analysis.} This figure shows the trend of FVD scores during long-term video generation. Compared to the baseline, GF results in significantly lower FVD after 100 frames.}
        \label{fig:accumulation_error}
    \end{minipage}
\end{figure}

As shown in Fig.~\ref{fig:alignment_fvd_depth}, we also explore applying alignment at different layers of the video diffusion model~\citep{dfot}, which uses a 7-layer U-ViT~\citep{bao2023uvit} backbone (3 downsampling layers, 1 bottleneck layer, 3 upsampling layers). Aligning at layer 3 yields the best FVD-256 score while preserving FVD-16 performance.
\vspace{-0.5em}
\paragraph{Mitigating Exposure Bias in Autoregressive Video Diffusion Model via Geometry Forcing.}

\vspace{-0.5em}
Exposure bias is a long-standing challenge in autoregressive video generation \citep{dforcing,dfot,ardiff,cheng2025nfd,huang2025selforcing}. While previous methods attempted to address it through memory mechanisms or context guidance, GF offers an orthogonal solution. As shown in Fig.~\ref{fig:accumulation_error}, GF mitigates long-term drift and reduces the accumulation of error during generation significantly by aligning 3D geometric representation. These results validate integrating 3D representation enables more reliable and coherent long-term video synthesis.

\subsection{User Study}
\begin{table}[t]
\centering
\caption{\textbf{User study.} Average scores on Camera Following, Object Consistency, and Scene Continuity. Each user has to rate each dimension on a scale of 1 to 5. Higher values indicate better quality.}
\vspace{-0.5em}
\begin{tabular}{lccc}
\toprule
\textbf{Method} & Camera Following & Object Consistency & Scene Continuity \\
\midrule
DFoT & 3.56 & 2.73 & 2.74 \\
REPA & 3.82 & 3.55 & 3.66 \\
VideoREPA & 3.31 & 3.05 & 2.82 \\
Geometry Forcing & \textbf{4.40} & \textbf{4.44} & \textbf{4.52} \\
\bottomrule
\end{tabular}
\label{tab:user_study}
\end{table}

\vspace{-0.5em}
While Reprojection Error (RPE) and Revisit Error (RVE) provide useful signals for measuring 3D consistency, they only capture specific geometric aspects and may miss perceptual artifacts or unrealistic dynamics that humans can easily notice. We therefore conduct a user study focusing on three aspects of 3D consistency. 1) \textbf{Camera Following}: Whether the camera in the video moves smoothly and accurately follows the given pose trajectory.  
2) \textbf{Object Consistency}: Whether objects remain consistent in shape, appearance, and position across frames.  
3) \textbf{Scene Continuity}: Whether the generated parts of the scene beyond the context frames remain coherent and reasonable.

We compare GF with DFoT~\citep{dfot}, REPA~\citep{yu2024representation}, and VideoREPA~\citep{zhang2025videorepa}. As shown in Tab.~\ref{tab:user_study}, GF consistently outperforms all baselines across the three aspects of 3D consistency, demonstrating its effectiveness in producing geometrically coherent videos.

\vspace{-0.7em}
\section{Conclusion}
\vspace{-0.5em}
This paper introduces Geometry Forcing (GF), a simple yet effective framework that enhances the geometric consistency of autoregressive video diffusion models by aligning their internal representations with geometry-aware features. Motivated by the observation that video diffusion models trained on raw pixel data often fail to capture meaningful 3D structure, our method introduces two alignment objectives, Angular Alignment and Scale Alignment, to guide latent features toward 3D-aware representations from a geometric foundation model. Empirical results on both camera-conditioned and action-conditioned video generation benchmarks demonstrate that GF significantly improves visual quality and 3D consistency, yielding lower FVD scores and more stable scene dynamics.

\textbf{Limitations.} The primary limitation of this work lies in its scale. While GF consistently improves geometric consistency and visual quality, its full potential remains unexplored under large-scale training. In particular, we have not yet investigated its effectiveness when applied to larger models and more extensive video datasets, which may further amplify its benefits.

\textbf{Future Work.} Future directions include scaling GF on larger datasets to build 3D-consistent world simulators, and applications for long video generation by treating 3D representation as memory.

\section*{Acknowledgment}

We would like to acknowledge Kiwhan Song and Boyuan Chen for their valuable advice and assistance in reproducing the Diffusion Forcing Transformer results.

\newpage

\section*{Reproducibility}

We provide comprehensive implementation details, including model architectures, training configurations, and data preprocessing procedures, in Appendix~\ref{supp:impl} to ensure reproducibility.

\bibliography{main}

@inproceedings{rombach2022high,
  title={High-resolution image synthesis with latent diffusion models},
  author={Rombach, Robin and Blattmann, Andreas and Lorenz, Dominik and Esser, Patrick and Ommer, Bj{\"o}rn},
  booktitle={Proceedings of the IEEE/CVF Conference on Computer Vision and Pattern Recognition (CVPR)},
  pages={10684--10695},
  year={2022}
}

@article{radford2019language,
  title={Language models are unsupervised multitask learners},
  author={Radford, Alec and Wu, Jeffrey and Child, Rewon and Luan, David and Amodei, Dario and Sutskever, Ilya and others},
  journal={OpenAI blog},
  volume={1},
  number={8},
  pages={9},
  year={2019}
}

@article{brown2020language,
  title={Language models are few-shot learners},
  author={Brown, Tom and Mann, Benjamin and Ryder, Nick and Subbiah, Melanie and Kaplan, Jared D and Dhariwal, Prafulla and Neelakantan, Arvind and Shyam, Pranav and Sastry, Girish and Askell, Amanda and others},
  journal={Advances in Neural Information Processing Systems (NeurIPS)},
  volume={33},
  pages={1877--1901},
  year={2020}
}

@article{ho2020denoising,
  title={Denoising diffusion probabilistic models},
  author={Ho, Jonathan and Jain, Ajay and Abbeel, Pieter},
  journal={Advances in Neural Information Processing Systems (NeurIPS)},
  volume={33},
  pages={6840--6851},
  year={2020}
}

@inproceedings{lipman2023flow,
  title={Flow Matching for Generative Modeling},
  author={Lipman, Yaron and Chen, Ricky TQ and Ben-Hamu, Heli and Nickel, Maximilian and Le, Matthew},
  booktitle={The Eleventh International Conference on Learning Representations},
  year={2023}
}

@inproceedings{liu2023flow,
  title={Flow Straight and Fast: Learning to Generate and Transfer Data with Rectified Flow},
  author={Liu, Xingchao and Gong, Chengyue and others},
  booktitle={The Eleventh International Conference on Learning Representations},
  year={2023}
}

@article{vaswani2017attention,
  title={Attention is all you need},
  author={Vaswani, Ashish and Shazeer, Noam and Parmar, Niki and Uszkoreit, Jakob and Jones, Llion and Gomez, Aidan N and Kaiser, {\L}ukasz and Polosukhin, Illia},
  journal={Advances in Neural Information Processing Systems (NeurIPS)},
  volume={30},
  year={2017}
}

@misc{sora,
  title={Sora},
  author={OpenAI},
  howpublished={\url{https://openai.com/index/sora/}},
  year=2024
}

@misc{veo3,
  title={Veo 3},
  author={Google},
  howpublished={\url{https://deepmind.google/models/veo/}},
  year={2025}
}

@article{videocrafter1,
  title={Videocrafter1: Open diffusion models for high-quality video generation},
  author={Chen, Haoxin and Xia, Menghan and He, Yingqing and Zhang, Yong and Cun, Xiaodong and Yang, Shaoshu and Xing, Jinbo and Liu, Yaofang and Chen, Qifeng and Wang, Xintao and others},
  journal={arXiv preprint arXiv:2310.19512},
  year={2023}
}

@inproceedings{kondratyuk2024videopoet,
  title={VideoPoet: A Large Language Model for Zero-Shot Video Generation},
  author={Kondratyuk, Dan and Yu, Lijun and Gu, Xiuye and Lezama, Jose and Huang, Jonathan and Schindler, Grant and Hornung, Rachel and Birodkar, Vighnesh and Yan, Jimmy and Chiu, Ming-Chang and others},
  booktitle={International Conference on Machine Learning},
  pages={25105--25124},
  year={2024},
  organization={PMLR}
}

@inproceedings{controlnet,
  title={Adding conditional control to text-to-image diffusion models},
  author={Zhang, Lvmin and Rao, Anyi and Agrawala, Maneesh},
  booktitle={Proceedings of the IEEE/CVF International Conference on Computer Vision},
  pages={3836--3847},
  year={2023}
}

@inproceedings{raj2023dreambooth3d,
  title={Dreambooth3d: Subject-driven text-to-3d generation},
  author={Raj, Amit and Kaza, Srinivas and Poole, Ben and Niemeyer, Michael and Ruiz, Nataniel and Mildenhall, Ben and Zada, Shiran and Aberman, Kfir and Rubinstein, Michael and Barron, Jonathan and others},
  booktitle={Proceedings of the IEEE/CVF international conference on computer vision},
  pages={2349--2359},
  year={2023}
}

@inproceedings{xiang2025TRELLIS,
  title={Structured 3d latents for scalable and versatile 3d generation},
  author={Xiang, Jianfeng and Lv, Zelong and Xu, Sicheng and Deng, Yu and Wang, Ruicheng and Zhang, Bowen and Chen, Dong and Tong, Xin and Yang, Jiaolong},
  booktitle={Proceedings of the Computer Vision and Pattern Recognition Conference},
  pages={21469--21480},
  year={2025}
}

@article{mildenhall2021nerf,
  title={Nerf: Representing scenes as neural radiance fields for view synthesis},
  author={Mildenhall, Ben and Srinivasan, Pratul P and Tancik, Matthew and Barron, Jonathan T and Ramamoorthi, Ravi and Ng, Ren},
  journal={Communications of the ACM},
  volume={65},
  number={1},
  pages={99--106},
  year={2021},
  publisher={ACM New York, NY, USA}
}

@article{kerbl20233d,
  title={3d gaussian splatting for real-time radiance field rendering},
  author={Kerbl, Bernhard and Kopanas, Georgios and Leimk{\"u}hler, Thomas and Drettakis, George},
  journal={ACM Transactions on Graphics},
  volume={42},
  number={4},
  pages={1--14},
  year={2023},
  publisher={ACM}
}

@article{heusel2017gans,
  title={Gans trained by a two time-scale update rule converge to a local nash equilibrium},
  author={Heusel, Martin and Ramsauer, Hubert and Unterthiner, Thomas and Nessler, Bernhard and Hochreiter, Sepp},
  journal={Advances in neural information processing systems},
  volume={30},
  year={2017}
}

@article{wang2004image,
  title={Image quality assessment: from error visibility to structural similarity},
  author={Wang, Zhou and Bovik, Alan C and Sheikh, Hamid R and Simoncelli, Eero P},
  journal={IEEE transactions on image processing},
  volume={13},
  number={4},
  pages={600--612},
  year={2004},
  publisher={IEEE}
}

@article{chen2024videocrafter2,
  title={Videocrafter2: Overcoming data limitations for high-quality video diffusion models},
  author={Chen, Haoxin and Zhang, Yong and Cun, Xiaodong and Xia, Menghan and Wang, Xintao and Weng, Chao and Shan, Ying},
  journal={arXiv preprint arXiv:2401.09047},
  year={2024}
}

@inproceedings{ranftl2021vision,
  title={Vision transformers for dense prediction},
  author={Ranftl, Ren{\'e} and Bochkovskiy, Alexey and Koltun, Vladlen},
  booktitle={Proceedings of the IEEE/CVF international conference on computer vision},
  pages={12179--12188},
  year={2021}
}

@inproceedings{peebles2023scalable,
  title={Scalable diffusion models with transformers},
  author={Peebles, William and Xie, Saining},
  booktitle={Proceedings of the IEEE/CVF international conference on computer vision},
  pages={4195--4205},
  year={2023}
}

@inproceedings{liu2025videodpo,
  title={Videodpo: Omni-preference alignment for video diffusion generation},
  author={Liu, Runtao and Wu, Haoyu and Zheng, Ziqiang and Wei, Chen and He, Yingqing and Pi, Renjie and Chen, Qifeng},
  booktitle={Proceedings of the Computer Vision and Pattern Recognition Conference},
  pages={8009--8019},
  year={2025}
}

@article{wan2025wan,
  title={Wan: Open and advanced large-scale video generative models},
  author={Wan, Team and Wang, Ang and Ai, Baole and Wen, Bin and Mao, Chaojie and Xie, Chen-Wei and Chen, Di and Yu, Feiwu and Zhao, Haiming and Yang, Jianxiao and others},
  journal={arXiv preprint arXiv:2503.20314},
  year={2025}
}

@InProceedings{Wu_2025_CVPR,
    author    = {Wu, Diankun and Liu, Fangfu and Hung, Yi-Hsin and Qian, Yue and Zhan, Xiaohang and Duan, Yueqi},
    title     = {4D-Fly: Fast 4D Reconstruction from a Single Monocular Video},
    booktitle = {Proceedings of the Computer Vision and Pattern Recognition Conference (CVPR)},
    month     = {June},
    year      = {2025},
    pages     = {16663-16673}
}

@article{li2024t2v,
  title={T2v-turbo: Breaking the quality bottleneck of video consistency model with mixed reward feedback},
  author={Li, Jiachen and Feng, Weixi and Fu, Tsu-Jui and Wang, Xinyi and Basu, Sugato and Chen, Wenhu and Wang, William Yang},
  journal={arXiv preprint arXiv:2405.18750},
  year={2024}
}

@article{yang2024cogvideox,
  title={Cogvideox: Text-to-video diffusion models with an expert transformer},
  author={Yang, Zhuoyi and Teng, Jiayan and Zheng, Wendi and Ding, Ming and Huang, Shiyu and Xu, Jiazheng and Yang, Yuanming and Hong, Wenyi and Zhang, Xiaohan and Feng, Guanyu and others},
  journal={arXiv preprint arXiv:2408.06072},
  year={2024}
}

@article{polyak2024movie,
  title={Movie gen: A cast of media foundation models. 2024a},
  author={Polyak, A and Zohar, A and Brown, A and Tjandra, A and Sinha, A and Lee, A and Vyas, A and Shi, B and Ma, CY and Chuang, CY and others},
  journal={arXiv preprint arXiv:2410.13720},
  year={2024}
}

@article{kong2024hunyuanvideo,
  title={Hunyuanvideo: A systematic framework for large video generative models},
  author={Kong, Weijie and Tian, Qi and Zhang, Zijian and Min, Rox and Dai, Zuozhuo and Zhou, Jin and Xiong, Jiangfeng and Li, Xin and Wu, Bo and Zhang, Jianwei and others},
  journal={arXiv preprint arXiv:2412.03603},
  year={2024}
}

@article{hu2023gaia,
  title={Gaia-1: A generative world model for autonomous driving},
  author={Hu, Anthony and Russell, Lloyd and Yeo, Hudson and Murez, Zak and Fedoseev, George and Kendall, Alex and Shotton, Jamie and Corrado, Gianluca},
  journal={arXiv preprint arXiv:2309.17080},
  year={2023}
}

@inproceedings{shin2024wham,
  title={Wham: Reconstructing world-grounded humans with accurate 3d motion},
  author={Shin, Soyong and Kim, Juyong and Halilaj, Eni and Black, Michael J},
  booktitle={Proceedings of the IEEE/CVF Conference on Computer Vision and Pattern Recognition},
  pages={2070--2080},
  year={2024}
}

@article{dforcing,
  title={Diffusion forcing: Next-token prediction meets full-sequence diffusion},
  author={Chen, Boyuan and Mart{\'\i} Mons{\'o}, Diego and Du, Yilun and Simchowitz, Max and Tedrake, Russ and Sitzmann, Vincent},
  journal={Advances in Neural Information Processing Systems},
  volume={37},
  pages={24081--24125},
  year={2024}
}

@article{guo2025mineworld,
  title={Mineworld: a real-time and open-source interactive world model on minecraft},
  author={Guo, Junliang and Ye, Yang and He, Tianyu and Wu, Haoyu and Jiang, Yushu and Pearce, Tim and Bian, Jiang},
  journal={arXiv preprint arXiv:2504.08388},
  year={2025}
}

@article{alonso2024atari,
  title={Diffusion for world modeling: Visual details matter in atari},
  author={Alonso, Eloi and Jelley, Adam and Micheli, Vincent and Kanervisto, Anssi and Storkey, Amos J and Pearce, Tim and Fleuret, Fran{\c{c}}ois},
  journal={Advances in Neural Information Processing Systems},
  volume={37},
  pages={58757--58791},
  year={2024}
}

@inproceedings{bruce2024genie,
  title={Genie: Generative interactive environments},
  author={Bruce, Jake and Dennis, Michael D and Edwards, Ashley and Parker-Holder, Jack and Shi, Yuge and Hughes, Edward and Lai, Matthew and Mavalankar, Aditi and Steigerwald, Richie and Apps, Chris and others},
  booktitle={Forty-first International Conference on Machine Learning},
  year={2024}
}

@article{valevski2024diffusion,
  title={Diffusion models are real-time game engines},
  author={Valevski, Dani and Leviathan, Yaniv and Arar, Moab and Fruchter, Shlomi},
  journal={arXiv preprint arXiv:2408.14837},
  year={2024}
}

@article{baker2022video,
  title={Video pretraining (vpt): Learning to act by watching unlabeled online videos},
  author={Baker, Bowen and Akkaya, Ilge and Zhokov, Peter and Huizinga, Joost and Tang, Jie and Ecoffet, Adrien and Houghton, Brandon and Sampedro, Raul and Clune, Jeff},
  journal={Advances in Neural Information Processing Systems},
  volume={35},
  pages={24639--24654},
  year={2022}
}

@article{teed2021droid,
  title={Droid-slam: Deep visual slam for monocular, stereo, and rgb-d cameras},
  author={Teed, Zachary and Deng, Jia},
  journal={Advances in neural information processing systems},
  volume={34},
  pages={16558--16569},
  year={2021}
}

@article{unterthiner2018towards,
  title={Towards accurate generative models of video: A new metric \& challenges},
  author={Unterthiner, Thomas and Van Steenkiste, Sjoerd and Kurach, Karol and Marinier, Raphael and Michalski, Marcin and Gelly, Sylvain},
  journal={arXiv preprint arXiv:1812.01717},
  year={2018}
}

@inproceedings{zhang2018unreasonable,
  title={The unreasonable effectiveness of deep features as a perceptual metric},
  author={Zhang, Richard and Isola, Phillip and Efros, Alexei A and Shechtman, Eli and Wang, Oliver},
  booktitle={Proceedings of the IEEE conference on computer vision and pattern recognition},
  pages={586--595},
  year={2018}
}

@article{ha2018world,
  title={World models},
  author={Ha, David and Schmidhuber, J{\"u}rgen},
  journal={arXiv preprint arXiv:1803.10122},
  year={2018}
}

@article{oasis2024,
  author    = {Decart and Julian, Quevedo and Quinn, McIntyre and Spruce, Campbell and Xinlei, Chen and Robert, Wachen},
  title     = {Oasis: A Universe in a Transformer},
  year      = {2024},
  url       = {https://oasis-model.github.io/}
}

@article{agarwal2025cosmos,
  title={Cosmos world foundation model platform for physical ai},
  author={Agarwal, Niket and Ali, Arslan and Bala, Maciej and Balaji, Yogesh and Barker, Erik and Cai, Tiffany and Chattopadhyay, Prithvijit and Chen, Yongxin and Cui, Yin and Ding, Yifan and others},
  journal={arXiv preprint arXiv:2501.03575},
  year={2025}
}

@book{glassner1989introduction,
  title={An introduction to ray tracing},
  author={Glassner, Andrew S},
  year={1989},
  publisher={Morgan Kaufmann}
}

@inproceedings{zhu2024compositional,
  title={Compositional 3D-aware Video Generation with LLM Director},
  author={Zhu, Hanxin and He, Tianyu and Tang, Anni and Guo, Junliang and Chen, Zhibo and Bian, Jiang},
  booktitle={The Thirty-eighth Annual Conference on Neural Information Processing Systems},
  year={2024}
}

@article{xu2024comp4d,
  title={Comp4d: Llm-guided compositional 4d scene generation},
  author={Xu, Dejia and Liang, Hanwen and Bhatt, Neel P and Hu, Hezhen and Liang, Hanxue and Plataniotis, Konstantinos N and Wang, Zhangyang},
  journal={arXiv preprint arXiv:2403.16993},
  year={2024}
}

@inproceedings{bahmani2024tc4d,
  title={Tc4d: Trajectory-conditioned text-to-4d generation},
  author={Bahmani, Sherwin and Liu, Xian and Yifan, Wang and Skorokhodov, Ivan and Rong, Victor and Liu, Ziwei and Liu, Xihui and Park, Jeong Joon and Tulyakov, Sergey and Wetzstein, Gordon and others},
  booktitle={European Conference on Computer Vision},
  pages={53--72},
  year={2024},
  organization={Springer}
}

@inproceedings{bar2025navigation_NVM,
  title={Navigation world models},
  author={Bar, Amir and Zhou, Gaoyue and Tran, Danny and Darrell, Trevor and LeCun, Yann},
  booktitle={Proceedings of the Computer Vision and Pattern Recognition Conference},
  pages={15791--15801},
  year={2025}
}

@inproceedings{niemeyer2021giraffe,
  title={Giraffe: Representing scenes as compositional generative neural feature fields},
  author={Niemeyer, Michael and Geiger, Andreas},
  booktitle={Proceedings of the IEEE/CVF conference on computer vision and pattern recognition},
  pages={11453--11464},
  year={2021}
}

@article{team2025aether,
  title={Aether: Geometric-aware unified world modeling},
  author={Aether and Zhu, Haoyi and Wang, Yifan and Zhou, Jianjun and Chang, Wenzheng and Zhou, Yang and Li, Zizun and Chen, Junyi and Shen, Chunhua and Pang, Jiangmiao and others},
  journal={arXiv preprint arXiv:2503.18945},
  year={2025}
}

@article{zhen2025tesseract,
  title={TesserAct: learning 4D embodied world models},
  author={Zhen, Haoyu and Sun, Qiao and Zhang, Hongxin and Li, Junyan and Zhou, Siyuan and Du, Yilun and Gan, Chuang},
  journal={arXiv preprint arXiv:2504.20995},
  year={2025}
}

@inproceedings{zhang2025world,
  title={World-consistent video diffusion with explicit 3d modeling},
  author={Zhang, Qihang and Zhai, Shuangfei and Martin, Miguel Angel Bautista and Miao, Kevin and Toshev, Alexander and Susskind, Joshua and Gu, Jiatao},
  booktitle={Proceedings of the Computer Vision and Pattern Recognition Conference},
  pages={21685--21695},
  year={2025}
}

@article{mai2025can,
  title={Can Video Diffusion Model Reconstruct 4D Geometry?},
  author={Mai, Jinjie and Zhu, Wenxuan and Liu, Haozhe and Li, Bing and Zheng, Cheng and Schmidhuber, J{\"u}rgen and Ghanem, Bernard},
  journal={arXiv preprint arXiv:2503.21082},
  year={2025}
}

@article{jiang2025geo4d,
  title={Geo4d: Leveraging video generators for geometric 4d scene reconstruction},
  author={Jiang, Zeren and Zheng, Chuanxia and Laina, Iro and Larlus, Diane and Vedaldi, Andrea},
  journal={arXiv preprint arXiv:2504.07961},
  year={2025}
}

@article{huang2025selforcing,
  title={Self Forcing: Bridging the Train-Test Gap in Autoregressive Video Diffusion},
  author={Huang, Xun and Li, Zhengqi and He, Guande and Zhou, Mingyuan and Shechtman, Eli},
  journal={arXiv preprint arXiv:2506.08009},
  year={2025}
}

@book{euler1845institutionum,
  title={Institutionum calculi integralis},
  author={Euler, Leonhard},
  volume={4},
  year={1845},
  publisher={impensis Academiae imperialis scientiarum}
}

@inproceedings{he2020momentum,
  title={Momentum contrast for unsupervised visual representation learning},
  author={He, Kaiming and Fan, Haoqi and Wu, Yuxin and Xie, Saining and Girshick, Ross},
  booktitle={Proceedings of the IEEE/CVF conference on computer vision and pattern recognition},
  pages={9729--9738},
  year={2020}
}

@article{williams1989learning,
  title={A learning algorithm for continually running fully recurrent neural networks},
  author={Williams, Ronald J and Zipser, David},
  journal={Neural computation},
  volume={1},
  number={2},
  pages={270--280},
  year={1989},
  publisher={MIT Press One Rogers Street, Cambridge, MA 02142-1209, USA journals-info~…}
}

@String(CVPR= {IEEE Conf. Comput. Vis. Pattern Recog.})

@String(CVPR  = {CVPR})

@inproceedings{bao2023uvit,
  title={All are worth words: A vit backbone for diffusion models},
  author={Bao, Fan and Nie, Shen and Xue, Kaiwen and Cao, Yue and Li, Chongxuan and Su, Hang and Zhu, Jun},
  booktitle={Proceedings of the IEEE/CVF conference on computer vision and pattern recognition},
  pages={22669--22679},
  year={2023}
}

@article{dfot,
  title={History-Guided Video Diffusion},
  author={Song, Kiwhan and Chen, Boyuan and Simchowitz, Max and Du, Yilun and Tedrake, Russ and Sitzmann, Vincent},
  journal={arXiv preprint arXiv:2502.06764},
  year={2025}
}

@article{mambaworldmodel,
  title={Long-context state-space video world models},
  author={Po, Ryan and Nitzan, Yotam and Zhang, Richard and Chen, Berlin and Dao, Tri and Shechtman, Eli and Wetzstein, Gordon and Huang, Xun},
  journal={arXiv preprint arXiv:2505.20171},
  year={2025}
}

@article{fuest2025maskflow,
  title={Maskflow: Discrete flows for flexible and efficient long video generation},
  author={Fuest, Michael and Hu, Vincent Tao and Ommer, Bj{\"o}rn},
  journal={arXiv preprint arXiv:2502.11234},
  year={2025}
}

@inproceedings{ardiff,
  title={Ar-diffusion: Asynchronous video generation with auto-regressive diffusion},
  author={Sun, Mingzhen and Wang, Weining and Li, Gen and Liu, Jiawei and Sun, Jiahui and Feng, Wanquan and Lao, Shanshan and Zhou, SiYu and He, Qian and Liu, Jing},
  booktitle={Proceedings of the Computer Vision and Pattern Recognition Conference},
  pages={7364--7373},
  year={2025}
}

@article{cheng2025nfd,
  title={Playing with Transformer at 30+ FPS via Next-Frame Diffusion},
  author={Cheng, Xinle and He, Tianyu and Xu, Jiayi and Guo, Junliang and He, Di and Bian, Jiang},
  journal={arXiv preprint arXiv:2506.01380},
  year={2025}
}

@article{parker2024genie2,
  title={Genie 2: A large-scale foundation world model},
  author={Parker-Holder, J and Ball, P and Bruce, J and Dasagi, V and Holsheimer, K and Kaplanis, C and Moufarek, A and Scully, G and Shar, J and Shi, J and others},
  journal={URL: https://deepmind. google/discover/blog/genie-2-a-large-scale-foundation-world-model},
  year={2024}
}

@article{xiao2025worldmem,
  title={WORLDMEM: Long-term consistent world simulation with memory},
  author={Xiao, Zeqi and Lan, Yushi and Zhou, Yifan and Ouyang, Wenqi and Yang, Shuai and Zeng, Yanhong and Pan, Xingang},
  journal={arXiv preprint arXiv:2504.12369},
  year={2025}
}

@article{videospm,
  title={Video World Models with Long-term Spatial Memory},
  author={Wu, Tong and Yang, Shuai and Po, Ryan and Xu, Yinghao and Liu, Ziwei and Lin, Dahua and Wetzstein, Gordon},
  journal={arXiv preprint arXiv:2506.05284},
  year={2025}
}

@article{chen2025flexworld,
  title={FlexWorld: Progressively expanding 3D scenes for flexiable-view synthesis},
  author={Chen, Luxi and Zhou, Zihan and Zhao, Min and Wang, Yikai and Zhang, Ge and Huang, Wenhao and Sun, Hao and Wen, Ji-Rong and Li, Chongxuan},
  journal={arXiv preprint arXiv:2503.13265},
  year={2025}
}

@article{Yu2024ViewCrafterTV,
  title={Viewcrafter: Taming video diffusion models for high-fidelity novel view synthesis},
  author={Yu, Wangbo and Xing, Jinbo and Yuan, Li and Hu, Wenbo and Li, Xiaoyu and Huang, Zhipeng and Gao, Xiangjun and Wong, Tien-Tsin and Shan, Ying and Tian, Yonghong},
  journal={arXiv preprint arXiv:2409.02048},
  year={2024}
}

@inproceedings{4d-fy,
  title={4d-fy: Text-to-4d generation using hybrid score distillation sampling},
  author={Bahmani, Sherwin and Skorokhodov, Ivan and Rong, Victor and Wetzstein, Gordon and Guibas, Leonidas and Wonka, Peter and Tulyakov, Sergey and Park, Jeong Joon and Tagliasacchi, Andrea and Lindell, David B},
  booktitle={Proceedings of the IEEE/CVF Conference on Computer Vision and Pattern Recognition},
  pages={7996--8006},
  year={2024}
}

@article{chung2023luciddreamer,
  title={Luciddreamer: Domain-free generation of 3d gaussian splatting scenes},
  author={Chung, Jaeyoung and Lee, Suyoung and Nam, Hyeongjin and Lee, Jaerin and Lee, Kyoung Mu},
  journal={arXiv preprint arXiv:2311.13384},
  year={2023}
}

@article{duan2025worldscore,
  title={Worldscore: A unified evaluation benchmark for world generation},
  author={Duan, Haoyi and Yu, Hong-Xing and Chen, Sirui and Fei-Fei, Li and Wu, Jiajun},
  journal={arXiv preprint arXiv:2504.00983},
  year={2025}
}

@article{he2024cameractrl,
  title={Cameractrl: Enabling camera control for text-to-video generation},
  author={He, Hao and Xu, Yinghao and Guo, Yuwei and Wetzstein, Gordon and Dai, Bo and Li, Hongsheng and Yang, Ceyuan},
  journal={arXiv preprint arXiv:2404.02101},
  year={2024}
}

@article{feng2024matrix,
  title={The matrix: Infinite-horizon world generation with real-time moving control},
  author={Feng, Ruili and Zhang, Han and Yang, Zhantao and Xiao, Jie and Shu, Zhilei and Liu, Zhiheng and Zheng, Andy and Huang, Yukun and Liu, Yu and Zhang, Hongyang},
  journal={arXiv preprint arXiv:2412.03568},
  year={2024}
}

@article{lee2024vividdream,
  title={Vividdream: Generating 3d scene with ambient dynamics},
  author={Lee, Yao-Chih and Chen, Yi-Ting and Wang, Andrew and Liao, Ting-Hsuan and Feng, Brandon Y and Huang, Jia-Bin},
  journal={arXiv preprint arXiv:2405.20334},
  year={2024}
}

@inproceedings{yu2025wonderworld,
  title={Wonderworld: Interactive 3d scene generation from a single image},
  author={Yu, Hong-Xing and Duan, Haoyi and Herrmann, Charles and Freeman, William T and Wu, Jiajun},
  booktitle={Proceedings of the Computer Vision and Pattern Recognition Conference},
  pages={5916--5926},
  year={2025}
}

@inproceedings{li2024megasam,
  title = {MegaSaM: Accurate, Fast and Robust Structure and Motion from Casual Dynamic Videos},
  author = {Li, Zhengqi and Tucker, Richard and Cole, Forrester and Wang, Qianqian and Jin, Linyi and Ye, Vickie and Kanazawa, Angjoo and Holynski, Aleksander and Snavely, Noah},
  booktitle={Proceedings of the IEEE/CVF Conference on Computer Vision and Pattern Recognition},
year={2025}
}

@inproceedings{wang2024dust3r,
  title={Dust3r: Geometric 3d vision made easy},
  author={Wang, Shuzhe and Leroy, Vincent and Cabon, Yohann and Chidlovskii, Boris and Revaud, Jerome},
  booktitle={Proceedings of the IEEE/CVF Conference on Computer Vision and Pattern Recognition},
  pages={20697--20709},
  year={2024}
}

@inproceedings{wang2025cut3r,
    Author = {Qianqian Wang* and Yifei Zhang* and Aleksander Holynski and Alexei A. Efros and Angjoo Kanazawa},
    Title = {Continuous 3D Perception Model with Persistent State},
    Year = {2025},
    booktitle={Proceedings of the IEEE/CVF Conference on Computer Vision and Pattern Recognition (CVPR)},
    }

@inproceedings{wang2025vggt,
  title={VGGT: Visual Geometry Grounded Transformer},
  author={Wang, Jianyuan and Chen, Minghao and Karaev, Nikita and Vedaldi, Andrea and Rupprecht, Christian and Novotny, David},
  booktitle={Proceedings of the IEEE/CVF Conference on Computer Vision and Pattern Recognition},
  year={2025}
}

@inproceedings{piccinelli2024unidepth,
  title={UniDepth: Universal monocular metric depth estimation},
  author={Piccinelli, Luigi and Yang, Yung-Hsu and Sakaridis, Christos and Segu, Mattia and Li, Siyuan and Van Gool, Luc and Yu, Fisher},
  booktitle={Proceedings of the IEEE/CVF Conference on Computer Vision and Pattern Recognition},
  pages={10106--10116},
  year={2024}
}

@inproceedings{zhang2025flare,
  title={Flare: Feed-forward geometry, appearance and camera estimation from uncalibrated sparse views},
  author={Zhang, Shangzhan and Wang, Jianyuan and Xu, Yinghao and Xue, Nan and Rupprecht, Christian and Zhou, Xiaowei and Shen, Yujun and Wetzstein, Gordon},
  booktitle={Proceedings of the Computer Vision and Pattern Recognition Conference},
  pages={21936--21947},
  year={2025}
}

@article{yang2025fast3r,
  title={Fast3R: Towards 3D Reconstruction of 1000+ Images in One Forward Pass},
  author={Yang, Jianing and Sax, Alexander and Liang, Kevin J and Henaff, Mikael and Tang, Hao and Cao, Ang and Chai, Joyce and Meier, Franziska and Feiszli, Matt},
  journal={arXiv preprint arXiv:2501.13928},
  year={2025}
}

@article{smart2024splatt3r,
  title={Splatt3r: Zero-shot gaussian splatting from uncalibrated image pairs},
  author={Smart, Brandon and Zheng, Chuanxia and Laina, Iro and Prisacariu, Victor Adrian},
  journal={arXiv preprint arXiv:2408.13912},
  year={2024}
}

@article{fan2025vlm3r,
  title={VLM-3R: Vision-Language Models Augmented with Instruction-Aligned 3D Reconstruction},
  author={Fan, Zhiwen and Zhang, Jian and Li, Renjie and Zhang, Junge and Chen, Runjin and Hu, Hezhen and Wang, Kevin and Qu, Huaizhi and Wang, Dilin and Yan, Zhicheng and others},
  journal={arXiv preprint arXiv:2505.20279},
  year={2025}
}

@article{wu2025spatial,
  title={Spatial-mllm: Boosting mllm capabilities in visual-based spatial intelligence},
  author={Wu, Diankun and Liu, Fangfu and Hung, Yi-Hsin and Duan, Yueqi},
  journal={arXiv preprint arXiv:2505.23747},
  year={2025}
}

@article{huang2025mllms,
  title={MLLMs Need 3D-Aware Representation Supervision for Scene Understanding},
  author={Huang, Xiaohu and Wu, Jingjing and Xie, Qunyi and Han, Kai},
  journal={arXiv preprint arXiv:2506.01946},
  year={2025}
}

@article{zhou2018stereo,
  title={Stereo magnification: Learning view synthesis using multiplane images},
  author={Zhou, Tinghui and Tucker, Richard and Flynn, John and Fyffe, Graham and Snavely, Noah},
  journal={arXiv preprint arXiv:1805.09817},
  year={2018}
}

@article{yu2024representation,
  title={Representation alignment for generation: Training diffusion transformers is easier than you think},
  author={Yu, Sihyun and Kwak, Sangkyung and Jang, Huiwon and Jeong, Jongheon and Huang, Jonathan and Shin, Jinwoo and Xie, Saining},
  journal={arXiv preprint arXiv:2410.06940},
  year={2024}
}

@article{zhang2025videorepa,
  title={VideoREPA: Learning Physics for Video Generation through Relational Alignment with Foundation Models},
  author={Zhang, Xiangdong and Liao, Jiaqi and Zhang, Shaofeng and Meng, Fanqing and Wan, Xiangpeng and Yan, Junchi and Cheng, Yu},
  journal={arXiv preprint arXiv:2505.23656},
  year={2025}
}

@article{oquab2023dinov2,
  title={Dinov2: Learning robust visual features without supervision},
  author={Oquab, Maxime and Darcet, Timoth{\'e}e and Moutakanni, Th{\'e}o and Vo, Huy and Szafraniec, Marc and Khalidov, Vasil and Fernandez, Pierre and Haziza, Daniel and Massa, Francisco and El-Nouby, Alaaeldin and others},
  journal={arXiv preprint arXiv:2304.07193},
  year={2023}
}

@article{IGVsurvey,
  title={A survey of interactive generative video},
  author={Yu, Jiwen and Qin, Yiran and Che, Haoxuan and Liu, Quande and Wang, Xintao and Wan, Pengfei and Zhang, Di and Gai, Kun and Chen, Hao and Liu, Xihui},
  journal={arXiv preprint arXiv:2504.21853},
  year={2025}
}

@article{Fei2024Driv3RLD,
  title={Driv3R: Learning Dense 4D Reconstruction for Autonomous Driving},
  author={Xin Fei and Wenzhao Zheng and Yueqi Duan and Wei Zhan and Masayoshi Tomizuka and Kurt Keutzer and Jiwen Lu},
  journal={ArXiv},
  year={2024},
  volume={abs/2412.06777},
  url={https://api.semanticscholar.org/CorpusID:274610426}
}

@article{Maggio2025VGGTSLAMDR,
  title={VGGT-SLAM: Dense RGB SLAM Optimized on the SL(4) Manifold},
  author={Dominic Maggio and Hyungtae Lim and Luca Carlone},
  journal={ArXiv},
  year={2025},
  volume={abs/2505.12549},
  url={https://api.semanticscholar.org/CorpusID:278739766}
}

@inproceedings{liu2025slam3r,
  title={Slam3r: Real-time dense scene reconstruction from monocular rgb videos},
  author={Liu, Yuzheng and Dong, Siyan and Wang, Shuzhe and Yin, Yingda and Yang, Yanchao and Fan, Qingnan and Chen, Baoquan},
  booktitle={Proceedings of the Computer Vision and Pattern Recognition Conference},
  pages={16651--16662},
  year={2025}
}

@article{ye2025fast,
  title={Fast autoregressive video generation with diagonal decoding},
  author={Ye, Yang and Guo, Junliang and Wu, Haoyu and He, Tianyu and Pearce, Tim and Rashid, Tabish and Hofmann, Katja and Bian, Jiang},
  journal={arXiv preprint arXiv:2503.14070},
  year={2025}
}

@misc{wang2025pi3,
      title={$\pi^3$: Scalable Permutation-Equivariant Visual Geometry Learning},
      author={Yifan Wang and Jianjun Zhou and Haoyi Zhu and Wenzheng Chang and Yang Zhou and Zizun Li and Junyi Chen and Jiangmiao Pang and Chunhua Shen and Tong He},
      year={2025},
      eprint={2507.13347},
      archivePrefix={arXiv},
      primaryClass={cs.CV},
      url={https://arxiv.org/abs/2507.13347}, 
}

@misc{wang2025spatialvlargescalevideodataset,
      title={SpatialVID: A Large-Scale Video Dataset with Spatial Annotations}, 
      author={Jiahao Wang and Yufeng Yuan and Rujie Zheng and Youtian Lin and Jian Gao and Lin-Zhuo Chen and Yajie Bao and Yi Zhang and Chang Zeng and Yanxi Zhou and Xiaoxiao Long and Hao Zhu and Zhaoxiang Zhang and Xun Cao and Yao Yao},
      year={2025},
      eprint={2509.09676},
      archivePrefix={arXiv},
      primaryClass={cs.CV},
      url={https://arxiv.org/abs/2509.09676}, 
}
\bibliographystyle{iclr2026_conference}

\newpage
\appendix

% Uncomment when the appendix is long
\setcounter{page}{1}

\section*{\textbf{Appendix for ICLR 2026 submission \textit{Geometry Forcing}: Marrying Video Diffusion and 3D Representation for Consistent World Modeling}}
\addcontentsline{toc}{section}{Paper Appendix for \textit{Geometry Forcing}}

\startcontents[chapters]
\printcontents[chapters]{}{1}{}

\section{Declaration of LLM usage}

We used large language models (LLMs) to aid or polish writing. Details are described in the paper. The use is limited to language editing (grammar, spelling, and word choice), code formatting (e.g., adding comments to the code). All scientific ideas, analysis, and conclusions were conceived, validated, and interpreted independently by the authors. We gratefully acknowledge the assistance of large language models in our work.

\section{Limitations}

Our method's reliance on VGGT (trained mainly on static scenes) constrains performance in dynamic environments. Generalization to significant motion scenarios requires further research.

\section{Implementation Details}
\label{supp:impl}

\subsection{Dataset}

\paragraph{RealEstate10K~\citep{zhou2018stereo}.} 
This dataset contains camera poses for 10 million video frames, suitable for evaluating 3D consistency and camera navigation in generated videos. We use a resolution of 256 $\times$ 256 pixels.

\paragraph{Minecraft~\citep{baker2022video}.} 
This game dataset includes action annotations, enabling evaluation of video generation in dynamic environments with camera motion.

\paragraph{Alignment Projection.}

To maximize geometric information retention, we aggregate features from all transformer blocks of the VGGT backbone as alignment targets. For computational efficiency, we apply bilinear interpolation to reduce the spatial dimensions from the original resolution to a manageable 512$\times$512 size. 

The alignment is performed using a Conv3D-based projector that operates on the latent dimensions. To accommodate multi-layer and multi-target alignment scenarios, we initialize independent projectors for each feature layer and target representation. This design ensures effective dimensional compatibility between the U-ViT feature space and the target geometric representations while maintaining computational efficiency.

\subsection{Training}

\paragraph{Model Architecture.} We adopt a U-ViT~\citep{bao2023uvit} backbone for video generation, with geometric feature alignment integrated at the third transformer block.

\paragraph{Training Data.} The model is trained on 10,000 video clips sampled from the RealEstate10K training dataset, each comprising 16 consecutive frames.

\paragraph{Training Protocol.} Training proceeds for 2 epochs using a learning rate of $8 \times 10^{-6}$ and a global batch size of 40. The geometric alignment loss is combined with the standard diffusion training objective.

\subsection{Inference}

A key advantage of Geometry Forcing is its inference-time efficiency, which introduces no computational overhead during sampling. We demonstrate results using a DDIM sampler with 50 steps, though the approach is compatible with any standard diffusion sampling algorithm.

\subsection{Metrics}
\label{supp:reproj}

In this section, we present the detailed implementations of the Reprojection Error (RPE) and the Revisit Error (RVE).

\paragraph{Reprojection Error.}~ Reprojection error (RPE) is a widely used metric in visual SLAM to evaluate multi-view geometric consistency. Following~\citet{duan2025worldscore}, we utilize DROID-SLAM~\citep{teed2021droid} to reconstruct the scene. Specifically, DROID-SLAM first extracts corresponding features across frames and then refines camera poses ($G_t$) and per-pixel depth estimates ($d_t$) through its differentiable Dense Bundle Adjustment (DBA) optimization, enforcing optical flow constraints and achieving robust structure-from-motion. The reprojection error is then computed by measuring the average Euclidean distance between the projected and observed pixel locations of co-visible 3D points across multiple frames. Formally, RPE is defined as:
\begin{equation}
RE= \frac{1}{|\mathcal{V}|} \sum_{(i,j) \in \mathcal{V}} \left\| \mathbf{p}^{*}_{ij} - \Pi(\mathbf{P}_{ij}) \right\|_2,
\end{equation}
where $\mathcal{V}$ denotes the set of valid feature correspondences, $\mathbf{p}^{}{ij}$ is the observed pixel location in generated video frames, $\mathbf{P}{ij}$ represents the corresponding reconstructed 3D point derived from refined depths and camera poses, and $\Pi$ denotes the camera projection function. Lower RPE values indicate better 3D alignment, reduced spatial artifacts, and enhanced spatio-temporal stability, thereby effectively reflecting the overall geometric coherence and consistency of the generated videos.

\paragraph{Revisit Error.}~Revisit Error evaluates long-range temporal consistency under full camera rotation, inspired by the setup proposed in WorldMem~\citep{xiao2025worldmem}. For each of 100 randomly sampled RealEstate10K video clips, we extract the first frame and initial camera pose. A camera trajectory of 256 frames is then constructed by rotating the initial camera pose around the Y-axis. We assess revisit consistency by comparing the first and final frame using reconstruction FID (rFID)~\citep{heusel2017gans}. Larger discrepancies indicate greater geometric or appearance drift, suggesting weaker long-term 3D consistency.

\subsection{3D Reconstruction from Diffusion Features}

In this section, we provide a detailed overview of the 3D reconstruction process illustrated in Fig.~\ref{fig:head}(c).

\paragraph{Reconstruction using Geometry Forcing Features.}  
We extract features from the Geometry Forcing (GF) model and pass them through the depth prediction head of VGGT to obtain the predicted depth map.

\paragraph{Reconstruction using Diffusion Features.}  

Motivated by our linear probing experiments, we investigate the 3D reconstruction capability of intermediate features extracted from DFoT~\citep{dfot}. Specifically, we freeze the pretrained DFoT backbone and train a DPT head~\citep{ranftl2021vision} to regress depth maps from its intermediate representations. The target depth maps are provided by the VGGT model~\citep{wang2025vggt} and serve as ground-truth supervision. The DPT head adopts the same architecture as the depth prediction module used in VGGT but is trained from scratch. We optimize the DPT head for 2500 steps using a learning rate of \(1 \times 10^{-4}\) and a batch size of 4.

\section{Supplementary Experiments}

\subsection{Ablation on Teacher Model}

Geometry Forcing does not depend on a specific 3D foundation model but still requires the 3D foundation to be feed-forward and to support multi-image inputs, as required by online training. We conduct Geometry Forcing algorithm on $Pi^3$ model and also achieves significant improvement on video generation as shown in Tab.~\ref{tab:teacher_model}. 

\begin{table}[t]
\centering
\caption{
\textbf{Ablation study on teacher model.} Our method (Geometry Forcing) is compatitable with different teacher models including VGGT and Pi3. \textbf{Bold} values denote the best, and \underline{Underlined} values indicate the second best.  \textbf{*} indicates the method is conditioned on the first frame only.}
\label{tab:teacher_model}
\small
\resizebox{\textwidth}{!}{
\begin{tabular}{lccccccc}
\toprule
\textbf{Method} & \textbf{Frames} & \textbf{FVD↓} & \textbf{LPIPS↓} & \textbf{SSIM↑} & \textbf{PSNR↑} & \textbf{RPE↓} & \textbf{RVE↓} \\
\midrule
DFoT~\citep{dfot}      & 256  & 364  & 0.55 & \underline{0.36} & 11.40 & 0.3575    & \underline{297}    \\
Pi3 \cite{wang2025pi3}     & 256  & \underline{309}  & \underline{0.53} &  \textbf{0.38} & \underline{11.53} & \textbf{0.3171}    & 303    \\
Geometry Forcing (VGGT)     & 256  & \textbf{243}  & \textbf{0.51} & \textbf{0.38} & \textbf{11.87} & \underline{0.3337}   & \textbf{272}    \\

\bottomrule
\end{tabular}
}
\end{table}

\subsection{Explicit Geometry Control}

We provide a full evaluation comparison between explicit control and our Geometry Forcing in Tab.~\ref{tab:explicit_control}.

\begin{table}[t]
\centering
\caption{
\textbf{Ablation study on explicit and implicit geometry information.} Our method (Geometry Forcing) achieves the best performance across all metrics on the RealEstate10K dataset for long-term (256-Frame) video generation. \textbf{bold} values denote the best, and \underline{Underlined} values indicate the second best.  \textbf{*} indicates the method is conditioned on the first frame only.}
\label{tab:explicit_control}
\small
\resizebox{\textwidth}{!}{
\begin{tabular}{lccccccc}
\toprule
\textbf{Method} & \textbf{Frames} & \textbf{FVD↓} & \textbf{LPIPS↓} & \textbf{SSIM↑} & \textbf{PSNR↑} & \textbf{RPE↓} & \textbf{RVE↓} \\
\midrule
DFoT~\citep{dfot}      & 256  & 364  & 0.55 & 0.36 & 11.40 & \underline{0.3575}   & \underline{297}    \\
Explicit geometry     & 256  & \underline{280}  & \underline{0.52} & \underline{0.37} & \textbf{11.99} & 0.3792   & \underline{297}   \\
Geometry Forcing (ours)     & 256  & \textbf{243}  & \textbf{0.51} & \textbf{0.38} & \underline{11.87} & \textbf{0.3337}   & \textbf{272}    \\

\bottomrule
\end{tabular}
}
\end{table}

\subsection{Alignment Context Length}
Geometry Forcing feeds 16 frames into the VGGT model to extract a latent representation, then aligns the first 16 frames during training. We present ablation results for different alignment context lengths in Tab.~\ref{tab:alignment_context}.  The results indicate that when the alignment context length is longer, the 3D information is more complete, thus leading to better results.

\begin{table}[t]
\centering
\caption{
\textbf{Ablation study on GF alignment context length.} Geometry Forcing-n indicates n frames is used to extract VGGT feature during training. The results are evaluated on the RealEstate10K dataset for long-term (256-Frame) video generation. \textbf{bold} values denote the best, and \underline{Underlined} values indicate the second best.  \textbf{*} indicates the method is conditioned on the first frame only.}
\label{tab:alignment_context}
\small
\resizebox{\textwidth}{!}{
\begin{tabular}{lccccccc}
\toprule
\textbf{Method} & \textbf{Frames} & \textbf{FVD↓} & \textbf{LPIPS↓} & \textbf{SSIM↑} & \textbf{PSNR↑} & \textbf{RPE↓} & \textbf{RVE↓} \\
\midrule
DFoT~\citep{dfot}      & 256  & 364  & 0.55 & \underline{0.36} & 11.40 & 0.3575    & 297    \\
Geometry Forcing-4    & 256  & 261  & \underline{0.51} & \textbf{0.38} & \textbf{12.21} & 0.3451   & 297   \\
Geometry Forcing-8      & 256  & \underline{257}  & \textbf{0.50} & \textbf{0.38} & \underline{12.17} & \textbf{0.3062}   & \underline{284}   \\
Geometry Forcing-16 (default)     & 256  & \textbf{243}  & \underline{0.51} & \textbf{0.38} & 11.87 & \underline{0.3337}   & \textbf{272}    \\

\bottomrule
\end{tabular}
}
\end{table}

\subsection{Multiple Layer Alignment}
Due to the large number of possible layer combinations for the layers selected for alignment, we present results for aligning the last three layers of our diffusion model in Tab.~\ref{tab:multiple_layer}. However, an increasing number of layers to align doesn't lead to better performance. 

\begin{table}[t]
\centering
\caption{\textbf{Alignment on Multiple Layers.} 
Comparison of aligning VGGT features at the middle layer vs.\ the last three layers of the diffusion model using Geometry Forcing.}
\label{tab:multiple_layer}
\small
\resizebox{\textwidth}{!}{
\begin{tabular}{lccccccc}
\toprule
\textbf{Method} & \textbf{Frames} & \textbf{FVD↓} & \textbf{LPIPS↓} & \textbf{SSIM↑} & \textbf{PSNR↑} & \textbf{RPE↓} & \textbf{RVE↓} \\
\midrule
DFoT~\citep{dfot} & 256 & 364 & 0.55 & 0.36 & 11.40 & \underline{0.3575} & \underline{297} \\

Geometry Forcing (Last 3 Layers) & 256 & \underline{280} & \underline{0.52} & \underline{0.37} & \textbf{11.99} & 0.3792 & \underline{297} \\

Geometry Forcing (Mid) & 256 & \textbf{243} & \textbf{0.51} & \textbf{0.38} & \underline{11.87} & \textbf{0.3337} & \textbf{272} \\
\bottomrule
\end{tabular}
}
\end{table}

\subsection{Text-to-video generation}

We extend our Geometry Forcing method to general text-to-video generation tasks. Our model is trained on 2K videos from the~\cite{wang2025spatialvlargescalevideodataset}, which provide detailed scene and camera descriptions. Experimental results demonstrate that our approach achieves improvements across multiple evaluation dimensions, including visual aesthetics, motion smoothness, and motion quality, as detailed in Table~\ref{tab:text2video-gf}. These results indicate that Geometry Forcing can extend effectively to dynamic text-to-video training, even though VGGT itself is trained on static scenes.

\begin{table}[h]
\centering
\caption{\textbf{Evaluation on text-conditioned video generation.} Evaluation on text-conditioned video generation. We report aesthetic quality, imaging quality, and motion smoothness for Wan2.1-1.3B before and after applying Geometry Forcing.}
\small
\begin{tabular}{lccc}
\toprule
\textbf{Method} & \textbf{Aesthetic Quality↑}  & \textbf{Imaging Quality↑}   & \textbf{Motion Smoothness↑}    \\
\midrule
Wan2.1            & 0.58 & 0.56 & 0.98  \\
Wan2.1 + GF & \textbf{0.59} & \textbf{0.59} & \textbf{0.99} \\
\bottomrule
\end{tabular}
\label{tab:text2video-gf}
\end{table}

\section{Discussion}

\subsection{Computational Efficiency}

We perform detailed profiling of our method on an NVIDIA A800 GPU and report the execution time and floating-point operations (FLOPs) for different components of our model during training in Table~\ref{tab:efficiency_breakdown}. The VGGT Feature Alignment contributes an additional 52.5\% in execution time and 60.4\% in total FLOPs.  Although this alignment process increases the per-step computation compared to the base diffusion model, it significantly accelerates convergence, thereby reducing the overall training duration. For fine-tuning, our method requires only a few thousand steps and completes within hours, yielding substantial efficiency gains over full pre-training. Additionally, during inference, our method incurs no additional computational cost compared to other methods that use explicit or implicit memory.

\begin{table}[h]
    \centering
    \setlength{\tabcolsep}{10pt}      % Increase column spacing
    \caption{\textbf{Training Stage Profiling.} We report the execution time and floating-point operations (FLOPs) for different components during a training step.}
    \label{tab:efficiency_breakdown}
    \resizebox{\textwidth}{!}{
    \begin{tabular}{l c c c c}
        \toprule
        \multirow{2}{*}{\textbf{Pipeline Stage}} & \multicolumn{2}{c}{\textbf{Time}} & \multicolumn{2}{c}{\textbf{FLOPs}} \\
        \cmidrule(lr){2-3} \cmidrule(lr){4-5}
                                         & \textbf{Value (s)} & \textbf{Percentage (\%)} & \textbf{Value (T)} & \textbf{Percentage (\%)} \\
        \midrule
        \multicolumn{5}{l}{\textit{Forward (Frozen)}} \\
        \hspace{1em} VGGT Encoding       & 0.853              & 53.4\%              & 93.3              & 60.4\%              \\
        \multicolumn{5}{l}{\textit{Forward (Learnable)}} \\
        \hspace{1em} Projector           & 0.017             & 1.1\%               & 0.1             & 0.1\%               \\
        \hspace{1em}Diffusion Backbone  & 0.220             & 13.8 \%              & 17.7              & 11.5\%              \\
        \multicolumn{5}{l}{\textit{Backward (Learnable)}} \\
        \hspace{1em} Projector + Diffusion Backbone      & 0.506             & 31.7\%              & 43.5              & 28.1\%              \\
        \midrule
        \textbf{Total per Step}          & \textbf{1.597}     & \textbf{100.0\%}    & \textbf{154.6}      & \textbf{100.0\%}    \\
        \bottomrule
    \end{tabular}
    }
\end{table}

\begin{figure}[htbp]
    \centering
    \includegraphics[
        width=\linewidth,
        clip
    ]{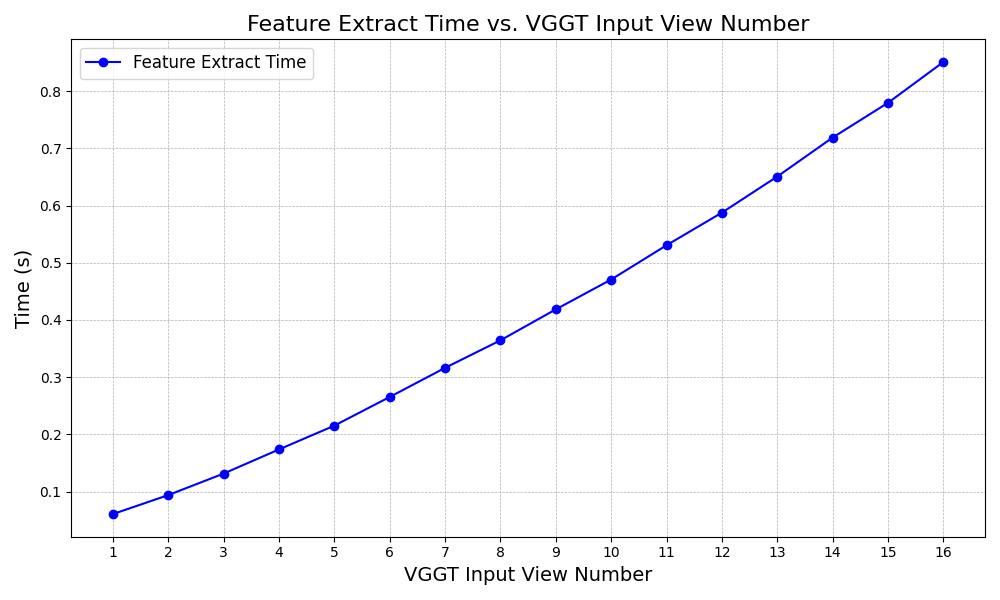}
    \caption{\textbf{VGGT Feature Extraction Time.} The feature extraction time of the VGGT model increases with the number of input views.}
    \label{fig:feature_extraction_time}
\end{figure}

We also provide a feature extraction time of the VGGT model in Fig.~\ref{fig:feature_extraction_time}. The result shows that the extraction time increases from 0.1s to 0.8s when the input increases from 1 to 12. 

\subsection{Analysis of Geometric and Semantic Representations}

We analyze the roles of geometric and semantic representation alignment in video generation. First, these representations exhibit considerable overlap rather than orthogonality. Semantic representations like DINOv2~\citep{oquab2023dinov2} demonstrate zero-shot depth estimation capabilities (see Section~7.5 and Figure~7 in the original paper), indicating inherent geometric understanding. Conversely, geometric representations such as VGGT utilize DINOv2 features as inputs, thereby encoding semantic information.

Second, experimental results in Table~\ref{tab:re10k_results} and Table~\ref{tab:target_representation} show that VGGT alignment primarily enhances 3D consistency, while DINOv2 alignment improves visual quality. The combination of both representations achieves superior performance compared to either individual approach.

Finally, the distinct contributions of each representation can be characterized as follows: semantic alignment enhances object realism and visual detail, whereas geometric alignment ensures structural consistency and shape coherence across generated video sequences.

\subsection{3D Consistency and Exposure Bias Mitigation}

As shown in Figure~\ref{fig:accumulation_error}, the FVD metric increases at a slower rate when Geometry Forcing is employed, indicating effective mitigation of exposure bias in long-term video generation. The underlying mechanism can be understood through the inherent stability of 3D scenes: while the number of generated frames increases, the underlying scene geometry remains the same. Geometry Forcing enables the model to internalize this geometric consistency, thereby reducing error accumulation when regenerating frames from previously encountered viewpoints.

\subsection{Failure Case Analysis}

\begin{figure}[htbp]
    \centering
    \includegraphics[
        width=\linewidth,
        clip
    ]{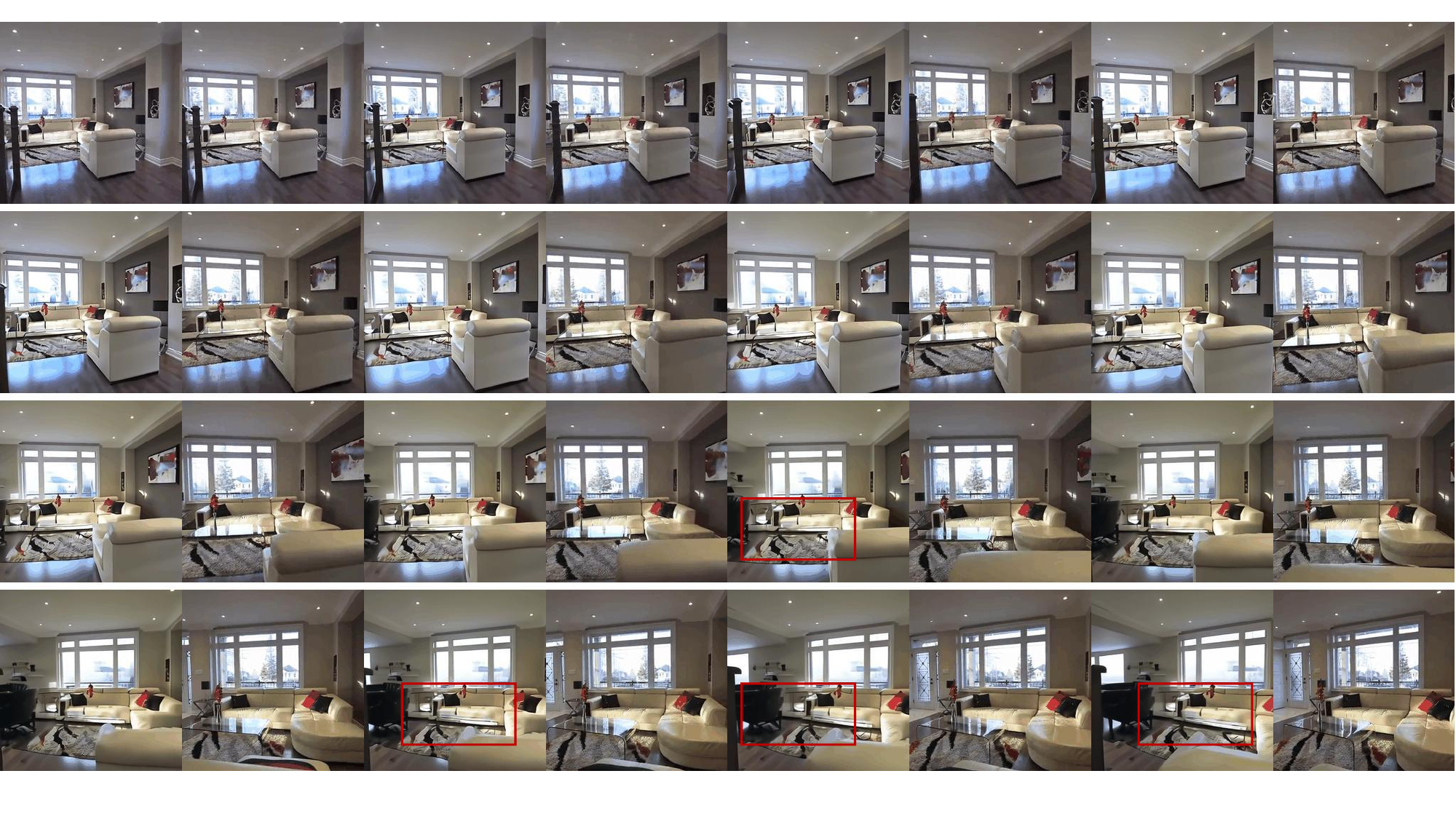}
    \caption{\textbf{Failure Case Analysis.} The transparent, reflective glass table intermittently disappears and reappears across frames, indicating that the model still has difficulty handling reflective materials. The red box indicates when the table disappears.}
    \label{fig:failure_case}
\end{figure}

Although our method significantly improves visual quality and geometric consistency in video generation, they still struggle in certain complex scenarios. As shown in Fig.~\ref{fig:failure_case}, the transparent, reflective glass table intermittently disappears and reappears across frames, indicating that the model still has difficulty handling reflective materials.

\newpage

\section{Supplementary Visualizations}
\label{supp:vis}
To better understand the effects of geometry, we provide comprehensive visual results.

\begin{figure}[htbp]
    \centering
    \includegraphics[
        width=\linewidth,
        clip
    ]{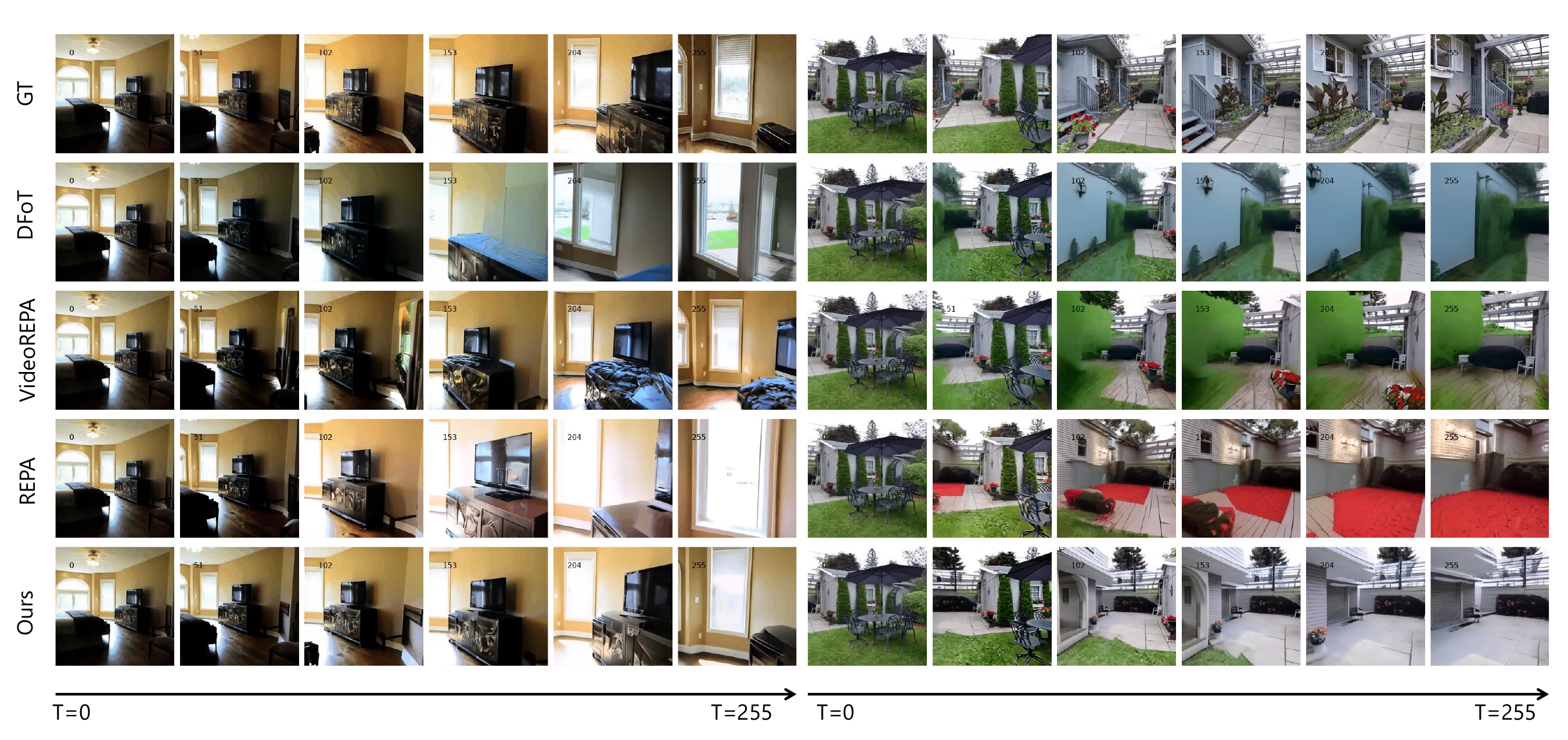}
    \caption{\textbf{Qualitative comparisons on camera-conditioned video generation.} All the videos are generated given the first frame and per-frame camera pose. We comprehensively compare GF (ours) with DFoT~\citep{dfot}, VideoREPA~\citep{zhang2025videorepa}, and REPA~\citep{yu2024representation}. The results demonstrate consistency in long-term video generation both inside (left) and outside (right) scenes.}
    \label{fig:main_comp}
\end{figure}

Fig.~\ref{fig:main_comp} presents qualitative comparisons on the RealEstate10K dataset. Given the same first frame and per-frame camera trajectory as input, we compare our proposed GF method with three strong baselines: DFoT~\citep{dfot}, REPA~\citep{yu2024representation}, and VideoREPA~\citep{zhang2025videorepa}. 

As shown in Fig.~\ref{fig:main_comp}, our method generates visually coherent and geometrically consistent videos over long time horizons, even with limited context. In particular, GF better preserves object shapes and scene layouts that are visible in context, while generating reasonable scenes not seen in the context. In contrast, baseline models often exhibit drift, shape distortion, or abrupt transitions. These results highlight the effectiveness of internalizing geometric priors to enhance spatial and temporal consistency in video generation.

\paragraph{Qualitative Ablation on Alignment loss.}

\begin{figure}[htbp]
    \centering
    \includegraphics[
        width=\linewidth,
        clip
    ]{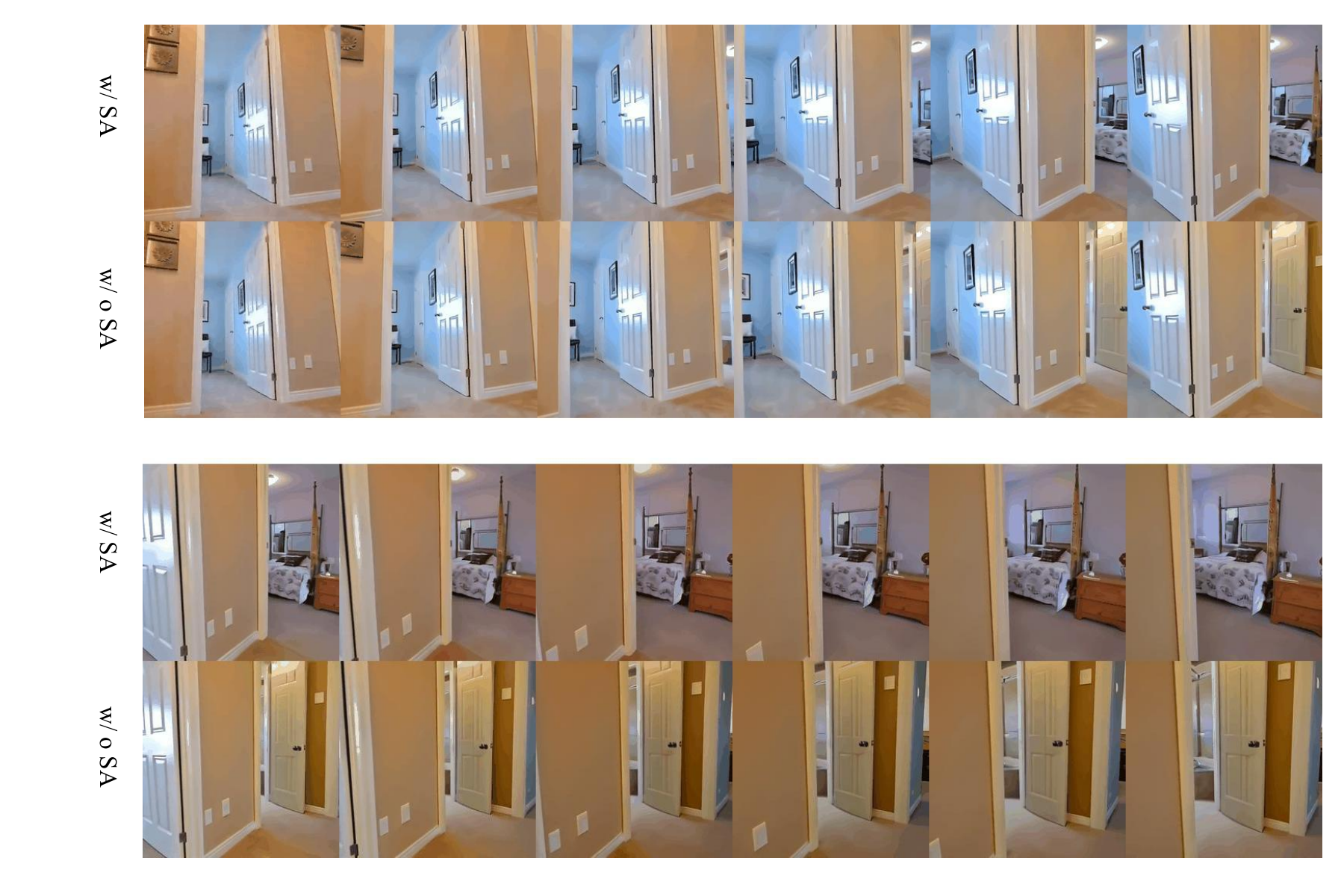}
    \caption{\textbf{Qualitative comparison of the Alignment Loss.} “w/ SA” denotes models trained with both angular alignment and scale alignment losses, while “w/o SA” refers to models trained using only angular alignment. Incorporating scale alignment enables the model to generate videos with more stable and realistic camera-following behavior.}
    \label{fig:scale_alignment}
\end{figure}

To further assess the impact of the proposed scale alignment loss, we conduct qualitative comparisons between models trained with and without this component (Fig.~\ref{fig:scale_alignment}). While angular alignment alone helps maintain basic geometric coherence, the lack of scale supervision often leads to inconsistent camera motion, manifesting as unstable perspective changes or unnatural object scaling. By introducing the scale alignment loss, our method produces noticeably smoother viewpoint transitions and more reliable camera-following behavior, demonstrating its effectiveness in stabilizing multi-frame geometry.

\end{document}